\documentclass[journal,letterpaper,10pt]{IEEEtran}
\usepackage{amsmath,amsfonts}
\usepackage{algorithm}
\usepackage{algorithmicx}%
\usepackage{algpseudocode}%
\usepackage{array}
\usepackage[caption=false,font=normalsize,labelfont=sf,textfont=sf]{subfig}
\usepackage{textcomp}
\usepackage{stfloats}
\usepackage{url}
\usepackage{verbatim}
\usepackage{graphicx}
\usepackage{xcolor}
\usepackage{cite}
\usepackage{amsmath}
\usepackage{multicol}
\usepackage{multirow}
\usepackage{listings}%
\usepackage{paralist}

\DeclareMathOperator{\rank}{rank}

\DeclareMathOperator{\mean}{mean}

\begin{document}

\title{Semi-Supervised Multi-View Crowd Counting by Ranking Multi-View Fusion Models}

\author{
\IEEEauthorblockN{Qi Zhang$^{1}$, Yunfei Gong$^{1}$, Zhidan Xie$^{1}$, Zhizi Wang$^{1}$, Antoni B. Chan$^{2}$, and Hui Huang*$^{1}$\IEEEcompsocitemizethanks{
    \IEEEcompsocthanksitem $^{1}$College of Computer Science and Software Engineering, Shenzhen University, Shenzhen, China. Email: qi.zhang.opt@gmail.com, \{gongyunfei2021,xiezhidan2021\}@email.szu.edu.cn, \{hongyeye114514,hhzhiyan\}@gmail.com
    \IEEEcompsocthanksitem $^{2}$Department of Computer Science, City University of Hong Kong, Hong Kong SAR, China. Email: abchan@cityu.edu.hk
    \IEEEcompsocthanksitem *Corresponding author.
}}}


\maketitle

\begin{abstract}
Multi-view crowd counting has been proposed to deal with the severe occlusion issue of crowd counting in large and wide scenes. However, due to the difficulty of collecting and annotating multi-view images, the datasets for multi-view counting have a limited number of multi-view frames and scenes. To solve the problem of limited data, one approach is to collect synthetic data to bypass the annotating step, while another is to propose semi-/weakly-supervised or unsupervised methods that demand less multi-view data. 
In this paper, we propose two semi-supervised multi-view crowd counting frameworks by ranking the multi-view fusion models of different numbers of input views, in terms of the model predictions or the model uncertainties.
Specifically, for the first method (\textit{vanilla} model), we rank the multi-view fusion models' prediction results of different numbers of camera-view inputs, namely, the model's predictions with fewer camera views shall not be larger than the predictions with more camera views. 
For the second method, we rank the estimated model uncertainties of the multi-view fusion models with a variable number of view inputs, guided by the multi-view fusion models' prediction errors, namely, the model uncertainties with more camera views shall not be larger than those with fewer camera views.
These constraints are introduced into the model training in a semi-supervised fashion for multi-view counting with limited labeled data. The experiments demonstrate the advantages of the proposed multi-view model ranking methods compared with other semi-supervised counting methods. 
\end{abstract}


\section{Introduction}
\IEEEPARstart{M}{ulti-view} crowd counting has been proposed to solve occlusion issues within large and complicated scenes for better counting performance. The training of multi-view crowd counting models requires a large number of multi-view images and labels. To extend the multi-view counting models to more practical applications, datasets with more scene and camera view variations are required. However, collecting and annotating real-world multi-view data is expensive and laborious.

To address the aforementioned issues, one possible approach is to generate synthetic data instead of collecting and annotating real-world scenes. The CVCS dataset \cite{zhang2021cross} is a large synthetic multi-view counting dataset, which consists of 31 scenes with 100 multi-view frames and around 100 camera views. Using a simple unsupervised domain transfer technique, a model trained on the CVCS dataset can be applied to real-world scenes well. The idea of using synthetic data for advancing task performance has also been adopted in the single-image counting task \cite{Wang2019Learning}.

Another possible approach to solve the issue of data scarcity is to develop semi-/weakly-supervised or unsupervised methods that rely on limited labeled data. For single-image counting, several semi-/weakly-supervised methods have been proposed to reduce the demand for the labeled crowd counting data and have achieved promising performance \cite{liu2019exploiting,liu2020semi,sam2019almost}. \cite{liu2018leveraging} leveraged the prediction ranking loss among image patches for better counting performance.
\cite{liu2020semi} constructed a series of surrogate tasks for the segmentation predictions on the image density maps as the self-training in the semi-supervised crowd counting. 
Considering that the aforementioned data scarcity issue is even more severe in the multi-view counting area, proposing semi-/weakly-supervised methods for multi-view crowd counting is a promising research topic. 
To the best of our knowledge, semi-/weakly-supervised methods has not yet been studied further in multi-view crowd counting research.
% \textcolor{blue}{To the best of our knowledge, systematic and dedicated exploration of semi-/weakly-supervised learning paradigms for multi-view crowd counting remains largely underexplored in existing literature.}

\begin{figure}[t]
\begin{center}
   \includegraphics[width=\linewidth]{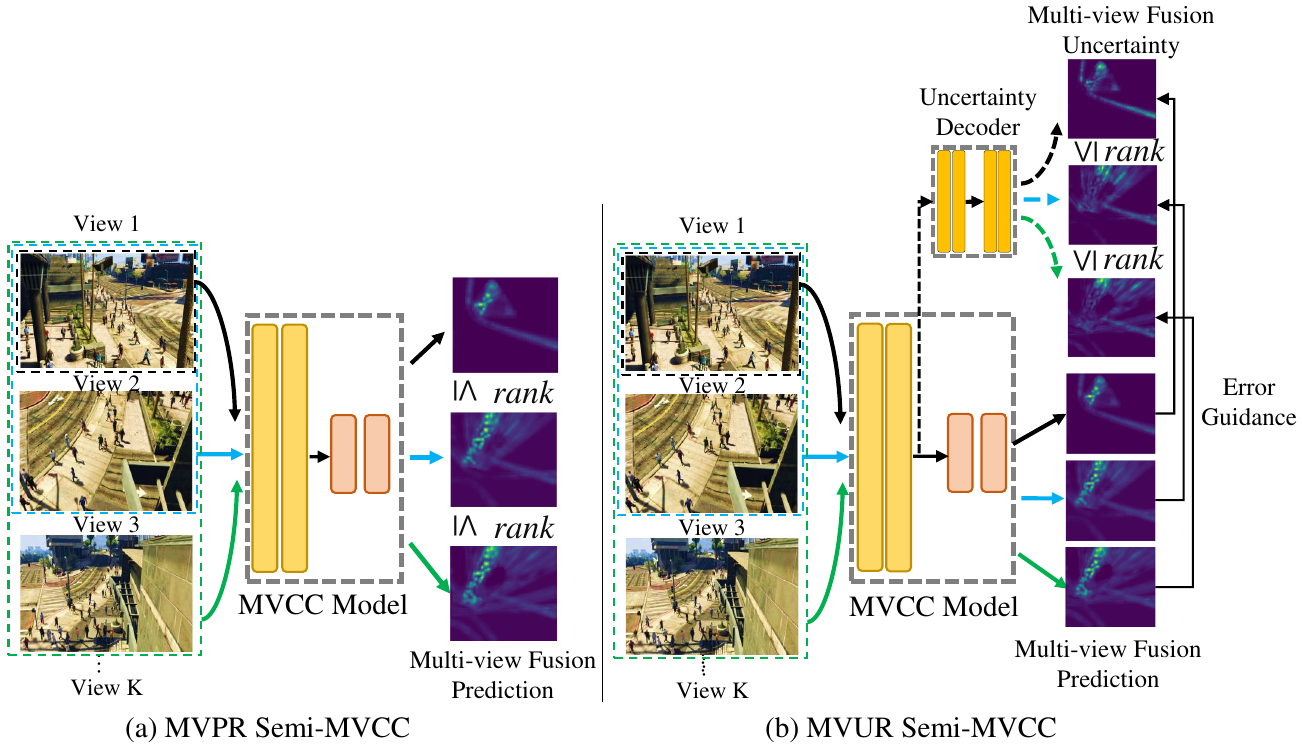}
\end{center}
\vspace{-0.4cm}
   \caption{The general pipeline of (a) our multi-view prediction ranking semi-supervised multi-view counting method (denoted as MVPR Semi-MVCC); and (b) our multi-view uncertainty ranking semi-supervised multi-view crowd counting method (denoted as MVUR Semi-MVCC): For the former, the multi-view fusion prediction results of fewer camera views shall not be larger  ($\leq$) than the counting results using more camera views. Additionally, for the latter, the estimated model uncertainties with more camera views shall not be larger ($\leq$) than those with fewer camera views. }
\vspace{-0.6cm}
\label{fig:demo_idea}
\end{figure}

In this paper, we propose two semi-supervised multi-view crowd counting frameworks via ranking the multi-view fusion models' predictions or uncertainties of different numbers of camera view inputs. The main idea of the frameworks is to introduce an extra constraint between multi-view fusion models of variable camera views in the semi-supervised model training.
Specifically, at first, for a good multi-view crowd counting model, the multi-view fusion count prediction of the same region on the scene ground map with fewer camera views should not be larger than those with more camera views, as more camera views can cover more people and generally handle occlusions better (see Fig. \ref{fig:demo_idea} (a), multi-view prediction ranking or MVPR). 
%Furthermore, we rank the predictions of different camera views of the same region and a loss is proposed to enforce that the ranking order is obeyed in the model training.
The proposed MVPR Semi-Supervised Multi-view Crowd Counting model consists of single-view feature extraction, multi-view feature projection and fusion, and multi-view decoding. The multi-view decoding module decodes inputs of various numbers of cameras, where a multi-view prediction ranking constraint (``$\leq$'') is enforced between the multi-view decoding results of different camera-view inputs for unlabeled examples.

\begin{figure*}[t]
\begin{center}
   \includegraphics[width=\linewidth]{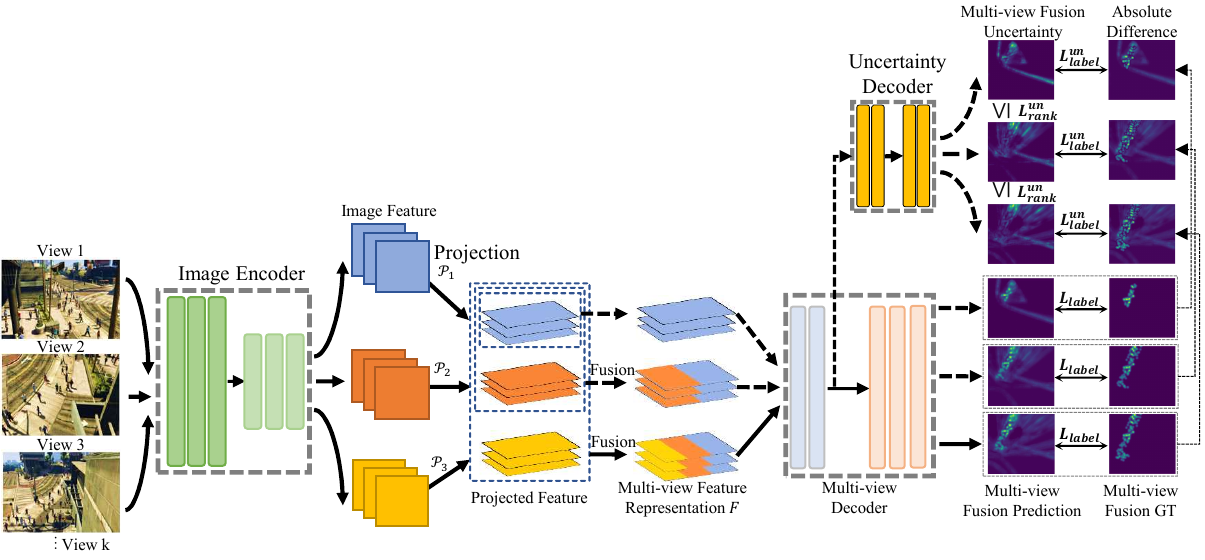}
\end{center}
\vspace{-0.3cm}
   \caption{The pipeline of the multi-view uncertainty ranking semi-supervised crowd counting method (MVUR Semi-MVCC). The proposed MVUR Semi-MVCC model uses the model uncertainty ranking instead of the model prediction ranking of different numbers of camera-view inputs for unlabeled data. The model comprises feature extraction of the image encoder, multi-view feature projection and fusion, model uncertainty decoder, and multi-view decoder. 
   Three losses are used in training: the multi-view fusion density map prediction loss $L_{label}$ and model uncertainty estimation loss $L^{un}_{label}$ for labeled data, and the multi-view fusion model uncertainty ranking loss $L^{un}_{rank}$ for unlabeled data. Dashed arrows refer to steps only used in the training.
   }
\vspace{-0.5cm}
\label{fig:MVUR}
\end{figure*}

The \textit{vanilla} model MVPR has a limitation that its effect is reduced when each view can cover most of the crowd in the scene and it may cause overcounting errors in the models. Therefore, to overcome the limitation of MVPR, we extend the idea from multi-view fusion model prediction ranking to model uncertainty ranking. Specifically, we first estimate the multi-view fusion models' uncertainties guided by the model prediction errors. We then rank the estimated model uncertainties according to the prior knowledge that the multi-view fusion models with more input camera views should have smaller uncertainties (``$\leq$'') compared to models with fewer camera views (see Fig. \ref{fig:MVUR}, multi-view uncertainty ranking or MVUR).
Furthermore, we rank the uncertainties of different camera views of the same region, and a loss is proposed to enforce that the ranking order is obeyed in the model training.
As in Fig. \ref{fig:MVUR}, compared to the MVPR model, the proposed MVUR Semi-MVCC model consists of an extra model uncertainty estimation network for different multi-view fusion models, which is supervised by each model's prediction errors. A model uncertainty ranking constraint is enforced between the multi-view fusion model uncertainty estimation network for unlabeled examples.

%Furthermore, we rank the predictions or uncertainties of different camera views of the same region and a loss is proposed to enforce that the ranking order is obeyed in the model training, respectively.
Note that the proposed %multi-view prediction and uncertainty ranking methods 
MVPR and MUPR have considered the core difference between multi-view counting and single-image counting, namely, the multi-view fusion problem, which is unique in multi-view counting compared to single-image counting tasks.
%, plus the the density map prediction mean squared error loss (MSE) for labeled examples.
%The model is trained with 2 losses: the density map prediction mean squared error loss (MSE) for labeled examples and the multi-view prediction ranking loss (rank) for unlabeled examples. 
%The model's training losses include the density map prediction loss (for labeled examples) and the multi-view prediction ranking loss (for unlabeled examples). 
The experiments show the proposed methods' advantages over previous semi-supervised methods originally designed for single-image counting. In summary, the main contributions of the paper are as follows.
%\begin{itemize}

\begin{itemize}
    \item To the best of our knowledge, this is the first study on semi-supervised multi-view crowd counting, which reduces the demand for annotated multi-view data in the multi-view crowd counting task.
    \item We consider the core difference between multi-view crowd counting and single-image crowd counting tasks, and propose a semi-supervised \textit{vanilla} method (MVPR Semi-MVCC) by ranking the multi-view fusion results of variable numbers of camera view inputs.
    \item We also propose MVUR, which avoids the limitations of MVPR and uses a more precise modeling method to enforce the ranking constraint between different multi-view fusion models. In addition, we propose a novel model uncertainty estimation network guided by the model prediction errors.   
    \item The experiments demonstrate that the proposed MVPR and MVUR Semi-MVCC methods perform better than existing semi-supervised counting methods originally designed for the single-image counting task.
\end{itemize}

\section{Related Work}
Crowd counting has explored various supervision levels and data modalities. Fully-supervised multi-view crowd counting methods, which leverage synchronized images and dense annotations, currently dominate due to their accuracy. However, multi-view data can naturally extend to semi-supervised approaches by utilizing its inherent geometric consistency to reduce label reliance. For single-image counting, researchers have already explored semi-, weakly-, and unsupervised methods, leveraging partial or no labels. Bridging multi-view and semi-supervised counting offers a promising direction to balance performance and annotation cost.

\subsection{Fully-supervised multi-view counting}

Existing multi-view counting methods rely on fully-supervised training, including both traditional and deep learning methods are as follows:
\textbf{Traditional} multi-view counting methods use background subtraction techniques and hand-crafted features, which can be further categorized into three types: detection \cite{viola2004robust, Sabzmeydani2007detecting}, regression \cite{chen2012feature, marana1998, chan2012counting},
and density map methods \cite{wang2016fast,xu2016crowd}.
These traditional methods are usually validated on the single-scene dataset PETS2009 \cite{pets2009} with limited frames and camera view variations, and their performances are limited by the weak feature representation power and the quality of the foreground/background extraction results.

\textbf{Deep learning} methods are based on designing end-to-end multi-view counting neural networks, and assume that the multi-cameras are synchronized and calibrated.
The deep-learning-based multi-view counting model generally consists of single-view feature extraction, feature projection, and multi-view fusion and decoding. A ground-plane density map is estimated to indicate the bird-view crowd density of all people in the scene.
\cite{zhang2019wide} proposed the first end-to-end DNN-based framework for multi-view crowd counting, and also introduced a multi-view dataset CityStreet. The work was extended to handle the rotation effects of projected features via a feature rotation selection module \cite{zhang2022wide}.
\cite{zhang2020_3d,zhang2022_3d} adopted the 3D feature fusion and the 3D density map supervision for solving the multi-view counting task in 3D space. Instead of being trained and tested on single scenes, \cite{zhang2021cross} proposed a large synthetic multi-view dataset CVCS and a camera selection fusion model for handling the cross-view cross-scene setting in real-world applications.
\cite{zheng2021learning} improved the late fusion model \cite{zhang2019wide} by adding the correlation between each pair of views. 
Recently, \cite{zhai2022co} proposed a graph-based multi-view learning model for multi-view counting. Besides multi-view counting, there are also methods for multi-view people detection that fuse multiple camera views for better people localization on the ground plane, such as \cite{hou2020multiview, hou2021multiview, song2021stacked,qiu20223d,hou2024learning}.
% zhang2024mahalanobis,zhang2024multi,
% \textcolor{black}{\cite{freelunch25} proposed a ``free lunch" strategy to improve performance through cross-modal alignment/density supervision without increasing inference cost.
% \cite{2025omnicount} enabled multi-category counting via semantic-geometric priors and requires category supervision to ensure accuracy of class-specific counting in complex scenarios.
% \cite{countingstackedobjects25} addressed the counting challenge of occluded stacked objects based on geometric estimation, relying on supervised depth/occupancy data to reconstruct spatial geometry.
CountFormer \cite{mo2024countformer} relied on the Transformer framework that converts image-level features into scene-level volume representations via camera encoding and attention-based feature lifting. Some recent work have advanced counting tasks via cross-modal alignment \cite{freelunch25}, semantic-geometric prior-based multi-category counting \cite{2025omnicount}, and geometric-estimation-driven occluded stacked object counting \cite{countingstackedobjects25}, respectively.

All existing multi-view counting methods adopt a fully supervised training fashion. Most methods are only trained on single scenes with limited generalization abilities. To further extend the multi-view counting task to more practical scenarios, some methods have also relied on synthetic data \cite{zhang2021cross}.
\textit{No existing work has explored semi-supervised multi-view counting methods, which is an important research topic for advancing the field into wider applications.}

\subsection{Semi-/Weakly-/Un-supervised single-image counting}
% Regarding single-image counting, most methods are fully supervised \cite{sam2017switching,onoro2016towards,bai2020adaptive}.
Regarding single-image counting, most methods are fully supervised \cite{sam2017switching,bai2020adaptive}.
Among them, many have focused on handling the scale variation and perspective change issues 
% \cite{li2018csrnet,Liu2019Context,yan2019perspective,yang2020reverse,Jiang_2020_CVPR}.
% Other research works explore different forms of supervision (e.g., regression methods or loss functions)  \cite{Ma2019BayesianLF,Wan_2021_CVPR,song2021rethinking,cheng2022rethinking}, more domains and modalities \cite{Wang2019Learning,Liu_2021_CVPR}, or general object counting \cite{yang2021class,Ranjan_2021_CVPR}.
\cite{Liu2019Context,yang2020reverse,Jiang_2020_CVPR}.
Other research works explore different forms of supervision (e.g., regression methods or loss functions)  \cite{Wan_2021_CVPR,song2021rethinking,cheng2022rethinking}, more domains and modalities \cite{Wang2019Learning,Liu_2021_CVPR}, or general object counting \cite{yang2021class,Ranjan_2021_CVPR}.

\textcolor{black}{To extend crowd counting’s application scenarios, weakly supervised \cite{liu2019exploiting, borstel2016gaussian, yangweakly, zhao2020active,liu2024weakly}, semi-supervised \cite{zhou2018crowd,sindagi2020learning,liu2020semi,wang2023semi,lin2023optimal,zhu2023multi,pelhan2024dave, DBLP:journals/tip/WeiQMHG23, LI_2023_ICCV, 9788041} and unsupervised \cite{sam2019almost} approaches have been proposed. \cite{liu2018leveraging} used patch prediction ranking loss for better performance. \cite{liu2020semi} designed surrogate segmentation tasks on density maps for self-training. \cite{DBLP:journals/tip/WeiQMHG23} adopted multi-density representations and kernel mean embedding, achieving SOTA via count consistency supervision. \cite{LI_2023_ICCV} calibrated uncertainty to enhance pseudo-label reliability, reaching SOTA in semi-supervised settings. \cite{meng2021spatial, lin2022semi} proposed teacher-student frameworks and agency-guided methods respectively. Ding et al. \cite{9788041} presented an unsupervised cross-domain framework with feature alignment. Recent works also explore LLMs/prompts for unsupervised/zero-shot counting \cite{liang2023crowdclip,ma2024clip,guo2024regressor}. \cite{yang2025taste} used unlabeled data to enhance pseudo-label quality and boost the model's generalization. P2R \cite{lin2025p2r} proposed a region-based supervision approach to address noise issues in pseudo-labels generated from unlabeled data for semi-supervised point-based crowd counting.}

\textit{It is important to note that all the semi-supervised counting methods mentioned above are designed especially for single-image counting tasks. To the best of our knowledge, there are no semi-supervised multi-view counting methods (Semi-MVCC) in the literature}. It is unclear whether directly adopting these methods for semi-supervised multi-view counting could achieve satisfying performance using multi-view data with limited labels. Our proposed multi-view fusion model ranking Semi-MVCC methods, MVPR and MVUR, consider the core difference between multi-view counting and single-image counting and achieve better performance than these semi-supervised methods originally designed for single-image counting methods. 
In particular, we consider the influence of different views on the multi-view fusion model, while the multi-view fusion process does not exist in single-image methods, and therefore it has not been studied by existing single-image semi-supervised methods.

\section{Multi-view Model Ranking Semi-MVCC}
In this paper, we propose ranking constraints between the multi-view counting models with different numbers of camera view inputs, in terms of model predictions and model uncertainties, and then enforce the constraint in the model training. The constraint is implemented through a ranking loss between different multi-view counting models (with different number of camera inputs) to enable better semi-supervised performance with limited training data. We follow MVMS \cite{zhang2019wide} to adopt the synchronized and calibrated multi-view setting in experiments, and the labeled training data is reduced to $P\%$ ($P=5, 10, 20$) for the semi-supervised training setting.
In the following sections, we first introduce the details of the adopted multi-view crowd counting network (MVCC). Then, we present the semi-supervised MVCC model with the multi-view model prediction ranking loss (MVPR) between models with different numbers of view inputs. Furthermore, considering the limitations of MVPR, we propose the multi-view model uncertainty ranking loss (MVUR).

\subsection{Multi-view crowd counting network (MVCC)}

The multi-view counting model \cite{zhang2019wide} we adopt consists of 3 stages, as shown in Fig. \ref{fig:MVUR}, the multi-view decoder and the previous stages:
single-view feature extraction, multi-view feature projection and fusion, and multi-view decoding. Each stage is explained as follows.
%%%

% \begin{figure*}[t]
% \begin{center}
%    \includegraphics[width=\linewidth]{figures/Figure_MVFR_pipeline.pdf}
% \end{center}
% \vspace{-0.5cm}
%    \caption{
%    The pipeline of the multi-view prediction ranking semi-supervised multi-view crowd counting method (MVPR Semi-MVCC). It comprises single-view feature extraction, multi-view feature projection and fusion, and multi-view decoding. The model's training losses include the density map prediction loss (for labeled examples, $L_{label}$) and the proposed multi-view prediction ranking loss (for unlabeled examples, $L^{pre}_{rank}$). Dashed arrows refer to steps only used in the training.}

% %
% \label{fig:MVPR}
% \end{figure*}

\textbf{Single-view feature extraction.}
The widely-used backbone models are used to extract the single-view features, such as VGG-Net \cite{simonyan2014very,li2018csrnet}.
To deal with the scale variation issue of the crowd, a three-scale feature pyramid is adopted, where the smaller-scale features are interpolated to the size of the largest-scale features and then they are concatenated together to form a multi-scale feature representation for each view.
%\textcolor{blue}{a three-scale feature pyramid is used, and the small-scale features are interpolated and adjusted to the maximum scale to form a multi-scale element representation for each view.}
The feature extraction network is shared across all views.
Denote the set of all input camera view images as ${\cal{V}}_k= \{V_0, V_1, ..., V_{k-1}\}={\{V_i\}}_{i\in C_k}$, where $V_i$ stands for view $i$, $k$ is the camera view input number and $C_k=\{0, 1,..., k-1\}$.
Correspondingly, the set of all camera views' features is ${\cal{F}}_k= \{F_i\}_{i\in C_k}$, where $F_i$ refers to the camera view $i$'s feature.

\textbf{Multi-view feature projection and fusion.}
The extracted single-view features are first projected from the image plane to the common scene (average height) plane for fusion via a spatial transformer net (STN) \cite{jaderberg2015spatial}.
The projected feature of view $i$ is denoted as $P(F_i)$ and the set of all camera views' project features is ${\{P(F_i)\}}_{i\in C_k}$.
Suppose $C_j$ is any non-empty subset of $C_k$, where $j\in \{1,2,...,k\}$ indicates the number of elements in the subset $C_j$.
In addition to fusing all camera views together from the whole camera view set $C_k$ (as in \cite{zhang2019wide,zhang2021cross}), we also fuse the camera views together in subsets $C_j$ to generate predictions from different numbers of input cameras, which is required for our ranking losses.
%fully utilize the supervisory information, where no extra annotation effort is required.
The multi-camera view features in any subset $C_j$ are concatenated together, and then the obtained multi-view feature representation is denoted as 
$F(C_j)=cat({\{P(F_i)\}}_{i\in C_j})$, where $j\in \{1,2,...,k\}$.
%The fusion results of any multi-view subset $C_j$ are of the same size since the output of the max-pooling operation is fixed and invariant to the camera view input number and order.

\textbf{Multi-view decoding.}
All fused projected features $F(C_j)$ are decoded to predict the scene-level density maps on the average height plane.
The corresponding predictions are denoted as $S_j=D_j(F(C_j)) \in R^{h*w}$, where  $D_j$ is a multi-view decoder to decode features $F(C_j)$, and $h$ and $w$ are density map size respectively. The multi-view decoders $\{D_j\}$ are not shared but have the same structure for different multi-view feature representations. 
%The detailed layer settings of the single-view feature extraction and multi-view decoding can be found in the supplemental.

For training the MVCC model with labels, the mean absolute error (MSE) loss between the predicted and ground-truth scene-level density maps ($S$ and $S^{gt}$, respectively) is used, i.e., the training loss for labeled examples is
 \begin{equation}	
    \begin{aligned}
	L_{label} & = \mathrm{MSE}(S_k, {S^{gt}_k}) + \lambda \sum_{j \in M}{\mathrm{MSE}(S_j, {S^{gt}_j})},
    \end{aligned} \label{eq:mse}
 \end{equation} 
where the first term is the main task for predicting the scene-level crowd density maps covered by all $k$ camera view inputs, and the second term is an auxiliary task, which is to predict crowd density maps covered by fewer camera views ($j<k$). $\lambda$ denotes the weight coefficient for balancing the auxiliary task loss term, and a fixed value of 0.001 is used in all experiments. $M$ is the set of camera view numbers in the auxiliary task, and $M = \{1,2,..., k-1\}$.

\textbf{Model layer settings.}
The layer settings of the MVCC model are presented in Table \ref{table:layer_setting} (a).
In terms of feature extraction, we adopted the VGG-16 feature extraction framework, and a feature pyramid (FPN) to extract features of different scales to handle scale-variation issues well. After projection, features of different views are concatenated to form a complete representation. As for the multi-view decoder, a 4-layer CNN decoder is used.

\subsection{Multi-view prediction ranking (MVPR)} \label{sec:MVPR}

\begin{figure}[t]
\begin{center}
   \includegraphics[width=1\linewidth]{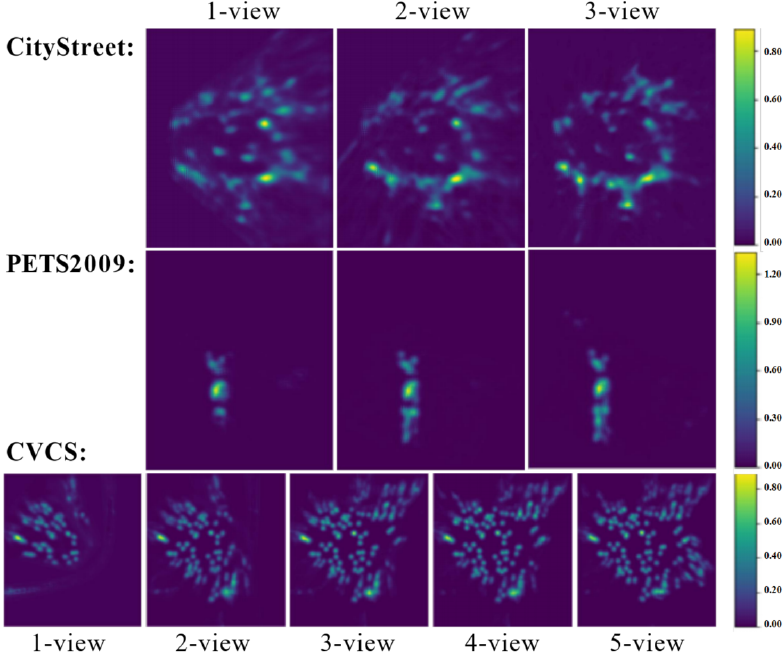}
\end{center}
\vspace{-0.5cm}
\caption{The density map predictions of labeled data using different numbers of camera views on the 3 datasets, which demonstrates the multi-view prediction ranking order between the fusion predictions of variable camera views.}
% \zq{decrease the dataset name font.}
% \yf{Done!}
% }
\vspace{-0.5cm}
\label{fig:ranking_vis}
\end{figure}

% When the labeled data is limited, only relying on the supervised training with (\ref{eq:mse}) will not be able to train the model well. Thus, we propose a semi-supervised multi-view counting method via the multi-view fusion model prediction ranking (MVPR) of variable views without any extra labels.
% The idea of MVPR is as follows: for an ideal multi-view crowd counting model, for a given region on the scene plane, the predicted count using a model with fewer camera views should not be larger than that of a model using more camera views.
% This is natural because more camera views can cover more people and provide more clues about the crowd in the scene. Specifically, using more camera views can handle the occlusion issue better than fewer camera views, which is the main advantage of multi-view crowd counting over single-image counting. Fig. \ref{fig:ranking_vis} presents the ranking order of multi-view fusion model predictions on 3 datasets, demonstrating that the fusion predictions of fewer camera views tend to be not larger than the predictions of more camera views.
\textcolor{black}{With limited labeled data, supervised training using Eq. (\ref{eq:mse}) alone fails to adequately train the model. We thus propose a \textit{vanilla} semi-supervised multi-view counting method via variable-view multi-view fusion model prediction ranking (MVPR), requiring no extra labels. The core idea of MVPR is that, for an ideal model, the predicted count of a given scene region using fewer camera views should not exceed that using more camera views. This is intuitive—more views cover more people, provide richer scene cues, and better handle occlusion (a key advantage of multi-view over single-view counting). Fig. \ref{fig:ranking_vis} visualizes these ranking results on 3 datasets, as the camera views increases, it's obvious that fewer-view fusion predictions tend to be no larger than more-view ones.}

\begin{table}
\footnotesize
% \begin{subtable}[c]{0.5\textwidth}
\centering
\begin{tabular}{|c|c|}
\hline
\multicolumn{2}{|c|}{Feature extractor} \\ \hline
Layer              & Filter      \\ \hline
conv 1             & $64\! \times\! 3\! \times\!  3\!  \times\!  3$, relu     \\ %\hline
conv 2             & $64\!  \times\!  64\!  \times\!  3\!  \times\!  3$, relu    \\ %\hline
pooling            & $2\!  \times\!  2\!  $         \\ %\hline
conv 3             & $128\!  \times\!  64\!  \times\!  3\!  \times\!  3$, relu   \\ %\hline
conv 4             & $128\!  \times\!  128\!  \times\!  3\!  \times\!  3$, relu \\ %\hline
pooling            & $2\!  \times\!  2\!  $          \\ %\hline
conv 5             & $256\!  \times\!  256\!  \times\!  3\!  \times\!  3$, relu\\ %\hline
conv 6             & $256\!  \times\!  256\!  \times\!  3\!  \times\!  3$, relu\\ %\hline
conv 7             & $256\!  \times\!  256\!  \times\!  3\!  \times\!  3$, relu \\ \hline

\multicolumn{2}{|c|}{Multi-view decoder}  \\ \hline
Layer & Filter     \\ \hline
conv 1     & $256\!  \times\!  256k\!  \times\!  3\!  \times\!  3$, relu   \\ %\hline
conv 2     & $128\!  \times\!  256\!  \times\!  3\!  \times\!  3$, relu   \\ %\hline
conv 3     & $64\!  \times\!  128\!  \times\!  3\!  \times\!  3$, relu   \\ %\hline
conv 4     & $1\!  \times\!  64\!  \times\!  1\!  \times\!  1$   \\ \hline

\multicolumn{2}{|c|}{Uncertainty estimation network} \\ \hline
Layer & Filter     \\ \hline
% conv 1     & $256\!  \times\!  256k\!  \times\!  1\!  \times\!  1$, relu   \\ %\hline
conv 1     & $128\!  \times\!  256\!  \times\!  3\!  \times\!  3$, relu   \\ %\hline
% conv 3     & $64\!  \times\!  128\!  \times\!  3\!  \times\!  3$, relu   \\ %\hline
conv 2     & $64\!  \times\!  128\!  \times\!  3\!  \times\!  3$, relu   \\ %\hline
conv 3     & $1\!  \times\!  64\!  \times\!  1\!  \times\!  1$   \\ \hline
\end{tabular}
%\vspace{-0.5cm}
% \caption {MVCC setting. In the first conv layer of the multi-view decoder, $k$ is the camera view input number.}
% \label{table:layer_setting}
% \centering
% \subcaption{MVCC setting.}
% \end{subtable}

% \begin{subtable}[c]{0.45\textwidth}
% \centering
% \begin{tabular}{|c|c|}
% \hline
% \multicolumn{2}{|c|}{Uncertainty estimation network.}  \\ \hline
% Layer & Filter     \\ \hline
% % conv 1     & $256\!  \times\!  256k\!  \times\!  1\!  \times\!  1$, relu   \\ %\hline
% conv 1     & $128\!  \times\!  256\!  \times\!  3\!  \times\!  3$, relu   \\ %\hline
% % conv 3     & $64\!  \times\!  128\!  \times\!  3\!  \times\!  3$, relu   \\ %\hline
% conv 2     & $64\!  \times\!  128\!  \times\!  3\!  \times\!  3$, relu   \\ %\hline
% conv 3     & $1\!  \times\!  64\!  \times\!  1\!  \times\!  1$   \\ \hline
% \end{tabular}
% \subcaption{Uncertainty estimation network setting. }
% \end{subtable}
\vspace{0.1cm}
\caption{The layer setting for the MVCC model and the Uncertainty estimation network in MVUR. In the first conv layer of the multi-view decoder, $k$ is the camera view input number.}
\label{table:layer_setting}
\vspace{-0.5cm}
\end{table}

Here we introduce this multi-view prediction ranking constraint on the \emph{unlabeled} data in the model training, and utilize it to improve the model performance under the semi-supervised training setting. Specifically, we require the density prediction of the same region under the scene map of fewer camera views ($S_{j_1}=D_{j_1}{(F(C_{j_1}))}$) is not larger than the prediction of the same region of more camera views ($S_{j_2}$), namely $S_{j_1} \leq S_{j_2} (j_1<j_2)$.
If this ranking constraint is violated by the multi-view counting model, a penalty should be enforced in the model, and thus a cost is obtained to force the model to obey the ranking order. To achieve this effect, a multi-view prediction ranking loss is proposed and introduced for unlabeled data as follows.
% \begin{equation}
\begin{align}
    L^{pre}_{rank} & =\sum_{j_1<j_2}{(\mean(\rank(S_{j_1}, {S_{j_2}})))} \\
                & = \sum_{j_1<j_2}{(\mean(\max(0, S_{j_1}-S_{j_2})))}.
\end{align}
% \begin{align}
%     L^{pre}_{rank} & =\sum_{j_1<j_2}{(\rank(S_{j_1}, {S_{j_2}}))} \\
%                 & = \sum_{j_1<j_2}{(\max(0, S_{j_1}-S_{j_2}))}.
% \end{align}
% \end{equation}
According to the ranking loss $L^{pre}_{rank}$, when density values in $S_{j_1}$ $\leq$ the corresponding density values in $S_{j_2}$, the loss gradient is 0;
When the ranking constraint is not obeyed by the MVCC model, for corresponding density values in $S_{j_1}>$ the density values in $S_{j_2}$, the ranking loss will penalize the density estimation $S_{j_1}$ and $S_{j_2}$ to be decreased and increased, respectively.
The choices of $(j_1, j_2)$ for different datasets are presented in the implementations.

\textbf{MVPR Semi-MVCC.}
Under the semi-supervised training setting, the labeled multi-view examples are not enough to train an effective multi-view crowd counting model.
To mitigate that, we introduce the multi-view fusion model prediction ranking on the unlabeled data in the model training, combining multi-view prediction ranking (MVPR) in the multi-view crowd counting (MVCC) model, denoted as MVPR Semi-MVCC. The whole training loss is rewritten as
 \begin{equation}	
    \begin{aligned}
	L & = L_{label} + \beta L^{pre}_{rank}, \\
    \end{aligned}
\label{eq:all_loss}
 \end{equation} 
where $\beta$ is the weight to balance the two terms in training. 

The proposed MVPR Semi-MVCC model is trained alternatively via the density map estimation loss and the multi-view prediction ranking loss. The MVCC network is first trained via the density map estimation loss $L_{label}$ with annotated multi-view images for several epochs to make full use of the labeled data. Then the model is trained by randomly and alternatively inputting labeled and unlabeled data.
Specifically, in each iteration, only labeled or unlabeled examples are input in the model, where the probability of randomly inputting labeled or unlabeled data is forced to be equal ($50\% : 50\%$). The labeled or unlabeled example is randomly selected from the labeled training set or the unlabeled training set, respectively. Thus, the labeled and unlabeled examples are randomly and alternatively input into the model.% for training.

\subsection{Multi-view uncertainty ranking (MVUR)} \label{sec:MVUR}
Our \textit{vanilla} model MVPR enforces that the model prediction resulting from more input camera views shall not be smaller than those with  with fewer views. However, this has a limitation that constraint is not effective when each view can see most of the views or introduces over-count errors in the model predictions with more views due to the ranking constraint. To avoid this weakness of MVPR, we propose the multi-view uncertainty ranking loss (MVUR) in the semi-MVCC model. 
The idea of MVUR is as follows: the multi-view models with more views tend to be more accurate than the models with fewer views, and thus the counting uncertainties of former models are smaller than the latter ones. Thus, we propose an extra model uncertainty estimation network supervised by the counting errors of the multi-view counting models with labeled data. Furthermore, estimated model uncertainties of different numbers of views are ranked as a constraint to train the semi-MVCC model.
Fig. \ref{fig:MVUR} presents the whole pipeline of the MVUR Semi-MVCC model: in addition to the similar modules as in MVPR, a new model uncertainty estimation network (see layer settings in Table \ref{table:layer_setting}) is introduced for MVUR and the ranking loss is enforced among model uncertainty estimations of different multi-view inputs. 

Specifically, the input of the proposed model uncertainty estimation network 
%for multi-view fusion models of different numbers of views as input 
is the multi-view feature representation $F'(C_j)$,
%of different camera views, 
where $C_j$ indicates the input view group. The uncertainty estimation process is denoted as 
 \begin{equation}	
    \begin{aligned}
    U_j = E(F'(C_j)),
    \end{aligned}
    \label{eq:uncertaintyNet}
 \end{equation} 
where $E$ represents the uncertainty estimation network, and $U_j$ is the estimated model uncertainty for the multi-view fusion models with $C_j$ as the input view group.

The supervision of the model uncertainty estimation network is based on the prediction error of the multi-view fusion model using the labeled data. The prediction error as $U^{gt}_j \in R^{h*w}$:
 \begin{equation}	
    \begin{aligned}
    U^{gt}_j = |S_j-{S^{gt}_j}|,
    \end{aligned}
    \label{eq:uncertaintyGT}
 \end{equation} 
where $|\cdot|$ is absolute value. Note that by definition of the ground truth of the model uncertainty, the corresponding predictions with lower uncertainty will also be more accurate.

An MSE loss between $U_j$ and $U^{gt}_j$ is used to train the model uncertainty estimation network for the labeled data, which is called the model uncertainty estimation loss $L^{un}_{label}$ :
 \begin{equation}	
    \begin{aligned}
    L^{un}_{label} = \mathrm{MSE}{(U_j, {U^{gt}_j})}.
    \end{aligned}
 \end{equation} \label{eq:uncertaintyLoss}

In MVUR, we introduce the multi-view fusion model uncertainty ranking constraint on the unlabeled data in the model training. Specifically, we require the estimated model uncertainty using more input views ($U_{j_2}$) to be not larger than that when %the estimated model uncertainty 
using fewer input views ($U_{j_1}$) on the common covered area. In other words,  $U_{j_2}\otimes M_{j_1} \leq U_{j_1} (j_1<j_2)\otimes M_{j_1}$, where $M_{j_1}$ represents the visible area mask under the view set of $j_1$. This means that when the model is provided with more camera views, the uncertainty of the resulting prediction should not be higher than when fewer views are used, forcing the MVCC models to produce more confident and accurate estimations with more views.  

If the ranking constraint is violated by the multi-view counting model -- i.e., if the uncertainty for more views is found to be greater than for fewer views -- a penalty is enforced in the model, denoted as the multi-view uncertainty ranking loss, 
\begin{align}
    L^{un}_{rank} & = \sum_{j_1<j_2}{\mean(\rank(U_{j_2}\otimes M_{j_1}, {U_{j_1}} \otimes M_{j_1}))} \\
                & = \sum_{j_1<j_2}{\mean(\max(0, U_{j_2} \otimes M_{j_1}-U_{j_1} \otimes M_{j_1}))},
\end{align}
% \begin{align}
%     L^{un}_{rank} & = \sum_{j_1<j_2}{\mean(\rank(U_{j_2}\otimes M_{j_1}, {U_{j_1}} \otimes M_{j_1}))} \\
%                 & = \sum_{j_1<j_2}{\mean(\max(0, U_{j_2} \otimes M_{j_1}-U_{j_1} \otimes M_{j_1}))},
% \end{align}
where $\otimes$ indicates element-wise multiply operation.
According to the ranking loss $L^{un}_{rank}$, in the common areas, where $U_{j_2}\otimes M_{j_1} \leq U_{j_1} \otimes M_{j_1}$, the loss gradient is 0;
where $U_{j_2}\otimes M_{j_1} > U_{j_1}\otimes M_{j_1}$, due to the uncertainty ranking constraint being violated, the loss is increased to penalize the model uncertainty estimation $U_{j_2}$ and $U_{j_1}$ to be decreased and increased. 
This penalty encourages the model to follow the expected ranking order. As a result, the model learns to improve its estimation and produces more reliable predictions on the unlabeled data.

\begin{figure}[t]
\begin{center}
   \includegraphics[width=1\linewidth]{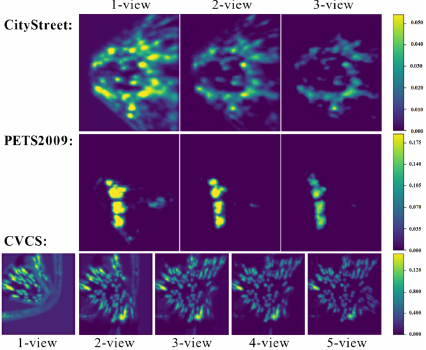}
\end{center}
\vspace{-0.5cm}
\caption{The predicted uncertainty maps when using different numbers of input camera views on the 3 datasets, which demonstrates their rank ordering.}
\vspace{-0.5cm}
\label{fig:ranking_uncertainty}
\end{figure}

Fig. \ref{fig:ranking_uncertainty} presents the multi-view fusion model uncertainty ranking order on 3 datasets, showing that the estimated model uncertainties using more camera views tend to be not larger than the uncertainties when using fewer camera views.

\textbf{MVUR Semi-MVCC.}
Combining multi-view fusion model uncertainty ranking (MVUR) in the multi-view crowd counting (MVCC) model, denoted as MVUR Semi-MVCC. The whole training loss is rewritten as
 \begin{equation}	
    \begin{aligned}
	L & = L_{label} + \eta L^{un}_{label} + \gamma L^{un}_{rank}, \\
    \end{aligned}
    \label{eq:all_loss2}
 \end{equation} 
where $\eta$ and $\gamma$ are the weights to balance the terms in training.
The training method is the same as in MVPR, where the labeled and unlabeled examples are input in the model randomly with equal probability.

\section{Experiments and Results}
\subsection{Experiment settings}

\textbf{Datasets.}
We introduce the multi-view crowd counting datasets used in the experiments: CVCS \cite{zhang2021cross}, CityStreet \cite{zhang2019wide} and PETS2009 \cite{pets2009}. For all datasets, $P\%$ ($P\in\{5, 10, 20\}$) of the multi-view frames in each training scene is selected as labeled data, and the remaining frames are used as unlabeled data in the semi-supervised training.
\textbf{CVCS} is a multi-view people dataset generated from a simulation.
    The dataset includes 23 training scenes and 8 testing scenes, where each scene includes 100 multi-view frames and about 100 camera views.
    The crowd number in each scene is 90-180.
    %The original image resolution is $ 1920 \times 1080 $, and the ground plane resolution $ 900 \times 800 $.
    % The camera calibration parameters are provided within the dataset.
    For training, 5 camera views are randomly selected for 5 times for each scene, while in testing 5 camera views are randomly selected for 21 times.
    The input image resolution is $640 \times 360$ and the ground plane resolution is $ 160 \times 180 $.
    %$P\% $ (P=5, 10 or 20) of the multi-view frames in each training scene is selected as labeled data and the rest frames are used as unlabeled data in the semi-supervised training.
\textbf{CityStreet} is a real single-scene multi-view dataset collected on the street intersection.
    The dataset contains 3 camera views, and 300 multi-view frames for training and 200 for testing.
    The crowd size ranges from 70-150.
    The input resolution is  $676 \times 380$ and the ground plane resolution is $ 160 \times 192 $.
    %$P\% $ (P=5, 10 or 20) of the multi-view frames in the training set is selected as labeled data and the rest training frames are used as unlabeled data in the semi-supervised training.
\textbf{PETS2009} is also a real single-scene multi-view dataset.
    The dataset contains 3 camera views, and 1105 multi-view frames are for training and 794 for testing.
    The crowd size ranges from 20-40.
    The input resolution is  $384 \times 288$ and the ground plane resolution is $ 152 \times 177 $.
    % $P\% $ (P=5, 10 or 20) of the multi-view frames in the training set is selected as labeled data and the rest training frames are used as unlabeled data in the semi-supervised training.
% \end{compactitem}
% We introduce the multi-view crowd counting datasets used in the experiments: CVCS \cite{zhang2021cross}, CityStreet \cite{zhang2019wide} and PETS2009 \cite{ferryman2009pets2009}.

\textbf{Evaluation metrics.}
We use the mean absolute error (MAE), mean square error (MSE), normalized (relative mean) absolute error (NAE) and grid average mean absolute error (GAME, denoted as $G(L)$, and $L$ is the grid dividing level) \cite{guerrero2015extremely} between the predicted counting number and the ground-truth crowd number as the evaluation metrics. Lower values indicate better multi-view counting performance in all metrics.
\begin{align}
    MAE = \frac{1}{N} \sum_{i}^{N} |c_i - {\hat{c}}_{i}|,\quad
%\end{align}
%\begin{align}
    MSE = \sqrt{\frac{1}{N} \sum_{i}^{N} (c_i - {\hat{c}}_{i})^2},
\end{align}
\begin{align}
    NAE = \frac{1}{N} \sum_{i}^{N} |c_i - {\hat{c}}_{i}|/{\hat{c}}_{i},
    \quad
%\end{align}
%\begin{align}
    G(L) = \frac{1}{N} \sum_{i}^{N} (\sum_{l=1}^{4^L} |c_i^l - {\hat{c}}_{i}^l|).
\end{align}
where $N$ is the number of the test images, $c_i$ and $\hat{c}_{i}$ are the estimated and ground-truth people count in the $i$-th image.
For the $GAME$ metric, the scene-level density maps are divided into $4^L$ non-overlapping patches, and the average $MAE$ of these patches is calculated.
$c_i^l$ and ${\hat{c}}_{i}^l$ are the estimated and ground-truth people count of the patch $l$ of $i$-th image. Note that $MAE$ equals $GAME$ with $L=0$.

\begin{table}[t]
\small
\centering
\begin{tabular}{@{\hspace{0.02cm}}l@{\hspace{0.08cm}}|l@{\hspace{0.08cm}}|c@{\hspace{0.08cm}}c@{\hspace{0.08cm}}c@{\hspace{0.08cm}}c@{\hspace{0.08cm}}c@{\hspace{0.08cm}}c@{\hspace{0.08cm}}c@{\hspace{0.08cm}}c@{\hspace{0.08cm}}c@{\hspace{0.08cm}}}
% \begin{tabular}{l|l|ccccc}
\hline
     & &  \multicolumn{5}{c}{CVCS}\\
% \hline
Semi-MVCC  & Rate   & MAE/G(0)  & G(1) & G(2)   & MSE  & NAE \\ \hline
\multirow{3}{*}{PatchRanking \cite{liu2018leveraging}}
&5\% & 10.26   & 16.29   & 25.43   & 13.15 & 0.094   \\
&10\% & 9.76   & \textbf{14.27}  & \textbf{22.14}  & 12.89 & {0.088}         \\
&20\% & 8.96   & 14.15   &\underline{21.49}   & 11.71 & 0.085      \\
\hline
\multirow{3}{*}{IRAST \cite{liu2020semi}}
&5\% & 10.53   & 16.19   & 24.56   & 14.07 & 0.096  \\
&10\% & 10.28   & 15.94   & 23.82   & 13.29 & 0.092    \\
&20\% & 9.23 & 14.88 & 23.13 & 11.90 & 0.084  \\
\hline
\multirow{3}{*}{DACount \cite{lin2022semi}}
&5\% & 10.24   & 15.05   & \underline{23.27}   & 13.46 & 0.098      \\
&10\% & 9.70  & 14.75   & 23.24   & {12.62} & 0.094         \\
&20\% & 8.71   & 13.94   & 22.91   &11.55  & 0.083      \\
\hline
\multirow{3}{*}{CountFormer \cite{mo2024countformer}}
&5\%  & 11.92   & 15.78   & 29.93    & 14.10  &  0.108     \\
&10\% & 10.88   & 15.06     & 29.12  & 13.62 &  0.106         \\
&20\% & 10.27   & 14.65    & 27.10    & 12.89  & 0.095       \\
\hline
\multirow{3}{*}{P2R \cite{lin2025p2r}}
&5\% &  33.30   &  - &  -  & 33.30  &  0.283   \\
&10\% & 18.94   & -  & -   &25.05   &  0.162  \\
&20\% & 17.42    & -  &  -  &24.78   &  0.146     \\
\hline
\multirow{3}{*}{Baseline1}
&5\% & 11.96   & 18.17   & 26.62   & 15.29 & 0.109    \\
&10\% & 10.87   & 16.27   & 25.27   & 13.84 & 0.100       \\
&20\% & 9.39   & 15.00   & 23.09   & 12.21 & 0.085    \\
\hline
\multirow{3}{*}{Baseline2}
&5\% & 11.75   & 17.91   & 26.12   & 15.29 & 0.105   \\
&10\% & 10.38   & 15.91   & 24.49   & 13.36 & 0.093   \\
&20\% & 9.22   & 14.62   & 22.59   & 12.16 & 0.084   \\
\hline
\hline
\multirow{3}{*}{MVPR (Vanilla) }
&5\%  &  \underline{9.79}   & \textbf{14.97}   & \textbf{23.19}   & \underline{12.82} & \underline{0.090}    \\
&10\% & \underline{9.36}   & \underline{14.49}   & \underline{22.18}   & \underline{12.17} &\underline{0.085}         \\
&20\% & \underline{8.64}   & \underline{13.86}   & 21.69   & \underline{10.98} & \underline{0.080}  \\
\hline
\multirow{3}{*}{MVUR (Ours) }
&5\% &  \textbf{9.73}   & \underline{15.04}   & 23.32  & \textbf{12.80} & \textbf{0.088}         \\
&10\% & \textbf{9.01}   & {14.78}   & 22.65   & \textbf{11.41} & \textbf{0.082}                  \\
&20\% & \textbf{8.09}   &\textbf{13.44}   & \textbf{21.42}   & \textbf{10.14} & \textbf{0.075}   \\
\hline
\end{tabular}
\vspace{0.1cm}
\caption{The comparison of the semi-supervised MVCC methods' performance on datasets CVCS under 5 metrics (the best result is in \textbf{bold} and the second best is \underline{underlined}).
Overall, the proposed MVUR and MVPR Semi-MVCC methods achieve the best and the second-best performance on the CVCS dataset, respectively.
}
\vspace{-0.5cm}
\label{table:CVCS_results}
\end{table}

\textbf{Comparison methods.}
\emph{Noting that there are no existing semi-supervised methods for multi-view crowd counting}, we thus adopted several single-image semi-supervised counting methods and modified them for the Semi-MVCC task.

\begin{itemize}

\item \textbf{PatchRanking Semi-MVCC.} The first comparison method is based on the PatchRanking \cite{liu2018leveraging}, denoted as PatchRanking Semi-MVCC. The comparison method uses a similar idea as \cite{liu2018leveraging} and introduces a patch ranking loss of regions on the predicted ground plane density maps, namely, the larger density map patches' sums are not smaller than the inner smaller patches' sums.

\item \textbf{IRAST Semi-MVCC.} 
The second comparison method is based on the IRAST \cite{liu2020semi}, denoted as IRAST Semi-MVCC. 
It adopts the same idea of IRAST \cite{liu2020semi}, which constructs a series of surrogate tasks based on the rule that the crowd in the high-threshold segmentation map must also exist in the low-threshold segmentation map after thresholding the ground-plane density map. The semi-supervised loss is employed to maintain consistency between the high-threshold and the low-threshold segmentation map based on the rule.

\item \textbf{DACount Semi-MVCC.}
The third comparison method is based on the DACount \cite{lin2022semi}, denoted as DACount Semi-MVCC. It adopts the idea of DACount \cite{lin2022semi} 
that the features within the same density level in the predicted ground plane density map should be as similar as possible, while foreground features should be distinct from background features. 
Building upon the original DACount model, we replaced the transformer network used for density map prediction with a regular CNN network. 
\textcolor{black}{\item \textbf{CountFormer.} 
Based on CountFormer \cite{mo2024countformer}, this method adopts its core multi-view volume aggregation module for fusing volumetric information into a scene-level 3D representation. Since the official implementation for CityStreet and PETS2009 is unavailable, we implemented CountFormer with the corresponding encoder and density map head from our model.}
\textcolor{black}{\item \textbf{P2R Semi-MVCC.} Derived from P2R \cite{lin2025p2r} , we retain its core loss calculation while adapting to multi-view scenarios. A multi-view feature aggregation module is added to project and concatenate cross-view features into a unified scene-level representation. Corresponding projection and preprocessing are applied to align with P2R’s original design.}

\item \textbf{Baseline.}
The baseline method is the Semi-MVCC model without using any multi-view fusion
model ranking modules. It only contains the main task loss ($\lambda=0$ and $\beta=0$)
or both the main task loss and auxiliary loss ($\lambda=0.001$ and $\beta=0$) in Eq. \ref{eq:all_loss}, denoted as \textbf{Baseline1} and \textbf{Baseline2}, respectively.

\end{itemize}

\textbf{Implementation details.}
In the model training, Adam is used, the batch size is 1, and the learning rate is 1e-5 for the whole training process. The input camera view order is random for all datasets. In the first 5 epochs, only the density map prediction loss on the label data is used to train the model. Later, in each iteration, the model is trained by inputting randomly-selected labeled data or unlabeled data alternatively. In the auxiliary density map prediction loss term of $L_{label}$, $M={1,2}$ for CityStreet and PETS2009, and  $M={1,2,3,4}$ for CVCS dataset. In the multi-view prediction ranking loss of $L^{un}_{rank}$, the $(j_1, j_2)$ combination is ${(1, 2), (2, 3)}$ for CityStreet and PETS2009, which means a local region's density prediction of 1 camera view is not larger than the prediction of 2 camera views, and 2 camera views' density prediction is not larger than the prediction of 3 camera views; the $(j_1, j_2)$ combination for CVCS dataset is ${(1, 2), (2, 3), (3, 4), (4, 5)}$. An Nvidia RTX3090 GPU is used to conduct all experiments. In all three datasets, we set $\beta=0.001$, $\eta=0.001$, $\gamma=0.001$.

% \textbf{Evaluation metrics.} We use the mean absolute error (MAE), mean square error (MSE), normalized (relative mean) absolute error (NAE), and grid average mean absolute error (GAME, denoted as $G(i)$, and $i$ is the grid dividing level) \cite{guerrero2015extremely} between the predicted counting number and the ground-truth crowd number as the evaluation metrics. Lower values indicate better multi-view counting performance in all metrics.

\begin{figure*}[t]
\begin{center}
   \includegraphics[width=0.8\linewidth]{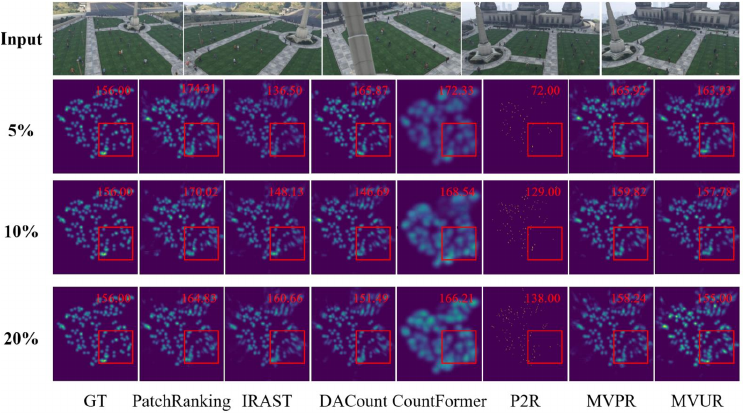}
\end{center}
\vspace{-0.5cm}
   \caption{The visualization results of the proposed MVPR and MVUR Semi-MVCC methods and the comparison methods on CVCS under different annotating rates. Overall, MVPR and MVUR Semi-MVCC achieve the best results among all methods.
   % \zq{The results of MVUR is not better under 5\% than IRAST. }
   }
\vspace{-0.5cm}
\label{fig:result_vis}
\end{figure*}

\subsection{Semi-supervised counting performance}

\textbf{CVCS.}
The performances of the proposed MVPR and MVUR Semi-MVCC methods and all comparison methods on the CVCS dataset are presented in Table \ref{table:CVCS_results}.
%According to Table \ref{table:CVCS_results}, 
The proposed MVPR and MVUR Semi-MVCC methods achieve the best results (MAE, MSE, and NAE) under all data annotating rates, which validates the advantage of their multi-view ranking constraints over the criteria from single-view ranking methods.
%method's advantage over comparison methods by introducing the in the model training.
Besides, MVUR achieves better performance than MVPR according to MAE, MSE, and NAE, which also demonstrates the advantage of model uncertainty ranking over model prediction ranking, because model uncertainty ranking avoids the issue of causing over-count errors in the model. Furthermore, when the data annotating rate increases from 5\% to 20\%, MVUR and MVPR achieve consistently better multi-view counting performance. This also demonstrates that the proposed multi-view ranking constraints can aid the model training with different labeling levels.  
\textcolor{black}{Due to practical implementation constraints, some comparison methods like CountFormer and P2R lag notably behind. Though CountFormer is a fully-supervised multi-view crowd counting method, and it does not consider utilizing unlabeled data. P2R is designed exclusively for single-image scenarios, lacking modules to fuse cross-view information or leverage inter-view complementary cues. Its single-view-focused supervision and limited metric support further hinder performance on complex multi-view benchmarks.}
Considering that the CVCS dataset is a large-scale cross-view cross-scene multi-view counting benchmark, the superior performance of our MVPR and MVUR on CVCS demonstrates well the multi-view prediction and uncertainty ranking constraint's effectiveness on complicated large scenes. 
The \textbf{visualization} results on the CVCS dataset are presented in Fig. \ref{fig:result_vis}. MVUR demonstrates superior performance compared to MVPR and comparison methods, particularly in the regions highlighted in red bounding boxes, where 
MVUR provides a clearer prediction of crowds in these areas, with a relatively smaller predicted count error.

\begin{table}[t]
\small
\centering
%\begin{tabular}{l@{\hspace{0.12cm}}|l@{\hspace{0.12cm}}|c@{\hspace{0.12cm}}c@{\hspace{0.12cm}}c@{\hspace{0.12cm}}|c@{\hspace{0.12cm}}c@{\hspace{0.12cm}}c@{\hspace{0.12cm}}
%|c@{\hspace{0.12cm}}c@{\hspace{0.12cm}}c@{\hspace{0.12cm}}}
% \begin{tabular}{l@{\hspace{0.2cm}}|l@{\hspace{0.2cm}}|c@{\hspace{0.2cm}}c@{\hspace{0.2cm}}c@{\hspace{0.2cm}}c@{\hspace{0.2cm}}c@{\hspace{0.2cm}}}
\begin{tabular}
{@{\hspace{0.02cm}}l@{\hspace{0.08cm}}|l@{\hspace{0.08cm}}|c@{\hspace{0.08cm}}c@{\hspace{0.08cm}}c@{\hspace{0.08cm}}c@{\hspace{0.08cm}}c@{\hspace{0.08cm}}c@{\hspace{0.08cm}}c@{\hspace{0.08cm}}c@{\hspace{0.08cm}}c@{\hspace{0.08cm}}}
\hline
     & &  \multicolumn{5}{c}{CityStreet}    \\
%\hline
Semi-MVCC  & Rate   & MAE/G(0)  & G(1) & G(2)   & MSE  & NAE     \\
\hline

\multirow{3}{*}{PatchRanking \cite{liu2018leveraging}}
&5\% & 8.04   & \underline{14.12}   & \underline{26.40}   & 9.85 & 0.103      \\
&10\% & 7.44  & 15.32   & 27.21  & 9.05 &0.095       \\
&20\% & 7.12 & 14.36   & 24.73   & 8.86 & 0.084     \\
\hline
\multirow{3}{*}{IRAST \cite{liu2020semi}}
&5\% & 8.00   & 15.47   & 27.55   & 9.79 & 0.103       \\
&10\% & \underline{7.36}   & 15.59   & 27.30   & \underline{9.08} & 0.091       \\
&20\% & 7.30   & 15.30   & 24.79   & 9.09 & 0.089        \\
\hline
\multirow{3}{*}{DACount \cite{lin2022semi}}
&5\% & 8.19   & 15.35   & 27.70   & 9.97 & 0.103     \\
&10\% & 7.78   & 16.84   & 27.42   & 9.53 & 0.097       \\
&20\% & \underline{6.95}   & 14.22   &23.68   & \underline{8.61}  & \underline{0.083}       \\
\hline
\multirow{3}{*}{CountFormer \cite{mo2024countformer}}
&5\% & 10.19 & 16.23   & 27.05    &12.47   & 0.112      \\
&10\% & 9.45   & 15.07   &  25.91   & 11.96  & 0.105          \\
&20\% & 8.41   & 14.52   & 24.13    & 11.58  & 0.089       \\
\hline
\multirow{3}{*}{P2R \cite{lin2025p2r}}
&5\% &12.36     & -  & -   &14.79   &0.160     \\
&10\% & 10.79   & -  & -   &13.13   &0.141     \\
&20\% & 10.08   & -  &  -  & 12.70  &   0.122  \\
\hline
\multirow{3}{*}{Baseline1}
&5\% & 10.41   & 18.61   & 29.40   & 12.89 & 0.123      \\
&10\% & 9.34   &\underline{15.11}   & 27.39   & 11.45 & 0.111       \\
&20\% & 8.23   & 13.77   & 25.88   & 10.37 & 0.097        \\
\hline
\multirow{3}{*}{Baseline2}
&5\% & 8.84   & 16.68   & 29.18   & 10.74 & 0.116      \\
&10\% & 8.45   & 16.76   & 29.09   & 10.22 & 0.112       \\
&20\% & 7.71   & 14.35   & \underline{23.64}   & 9.81 & 0.089        \\
\hline
\hline
\multirow{3}{*}{MVPR (Vanilla)}
&5\%  & \underline{7.99}  & 15.24   & 26.85   & \underline{9.96} & \underline{0.097}  \\
&10\% & 7.47  & 15.12   & \underline{26.60}  & 9.15 & \underline{0.090}       \\
&20\% & 7.28  & \underline{13.71}   & 24.29   & 8.93 & 0.092  \\
\hline
\multirow{3}{*}{MVUR (Ours)}
&5\%  & \textbf{7.79}   & \textbf{14.06} &\textbf{24.65} & \textbf{9.62} &\textbf{0.095}    \\
&10\% & \textbf{6.51}   & \textbf{14.33}   & \textbf{24.67}   & \textbf{8.57}    & \textbf{0.078}  \\
&20\% & \textbf{5.82}   & \textbf{12.93}   & \textbf{22.59}    & \textbf{7.77}  & \textbf{0.071}       \\
\hline
\end{tabular}
\vspace{0.1cm}
\caption{The comparison of the semi-supervised multi-view counting performance on CityStreet on 5 metrics (the best result is in \textbf{bold} and the second best is \underline{underlined}). Overall, MVUR Semi-MVCC achieves the best results. The proposed MVPR Semi-MVCC method achieves comparable performance to PatchRanking and IRAST Semi-MVCC, and is better than DACount Semi-MVCC and Baseline methods. 
%The possible reason is each camera view in CityStreet can cover most of the same crowd in the scene, which makes the multi-view prediction ranking constraint less effective.
}
\vspace{-0.5cm}
\label{table:CityStreet_results}
\end{table}

\begin{figure}[t]
\begin{center}
   \includegraphics[width=\linewidth]{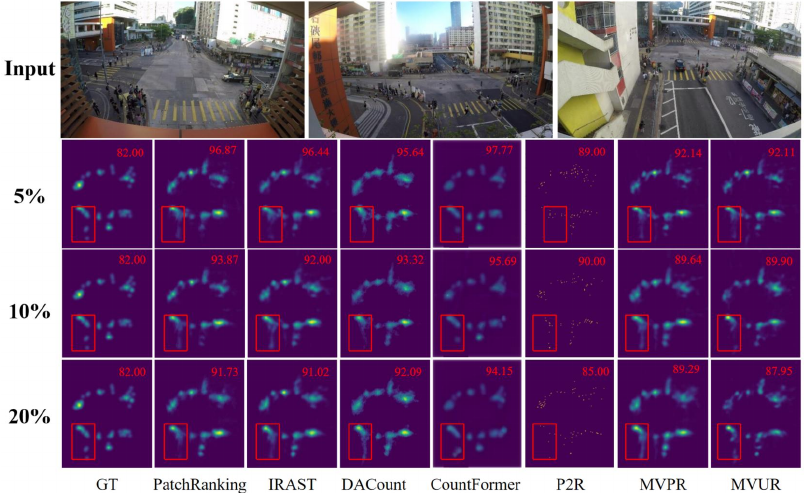}
\end{center}
%\vspace{-0.5cm}
   \caption{The visualization results of the proposed MVPR and MVUR Semi-MVCC method and the comparison methods on CityStreet under different annotating rates. 
   %The yellow, green, and blue lines in the `MVPR' represent the covered areas of different camera views on the ground plane. These lines indicate that each view of CityStreet captures most of the people in the scene, resulting in minimal improvement for MVPR compared to other methods.
   % \zq{make the methods name consistent with the text.} \yf{Done!}
}
%\vspace{-0.5cm}
\label{fig:CityStreet_vis}
\end{figure}

\textbf{CityStreet.}
The performances of the proposed MVPR and MVUR %Semi-MVCC 
and comparison methods on CityStreet are presented in Table \ref{table:CityStreet_results}. 
%The results reveal that
MVUR shows the best performance compared to all other methods. At all test rates (5\%, 10\%, 20\%), MVUR consistently achieves lower metric values. 
MVPR achieves the second according to MAE, MSE and NAE under the 5\% annotating rate, and is slightly worse than PatchRanking and IRAST under 10\% and 20\%. The possible reason why MVPR cannot perform quite well is that each camera view in the CityStreet dataset can cover most of the same crowd in the scene, thus the multi-view model prediction ranking constraint is not effective enough between predictions of different camera views (see Fig. \ref{fig:CityStreet_vis} MVPR).
%\zq{Show the camera views on the figure}. 
This shows a limitation of the proposed MVPR method: When each view captures the same crowd mostly, the effect of MVPR will be smaller. 
MVUR utilizes model uncertainty ranking constraint instead of model prediction ranking, which avoids the limitation in MVPR and thus achieves the best results.
The visualization results on CityStreet of all methods are presented in Fig. \ref{fig:CityStreet_vis}, and MVUR shows more accurate counting performance.

\begin{table}[t]
\small
\centering
%\begin{tabular}{l@{\hspace{0.12cm}}|l@{\hspace{0.12cm}}|c@{\hspace{0.12cm}}c@{\hspace{0.12cm}}c@{\hspace{0.12cm}}|c@{\hspace{0.12cm}}c@{\hspace{0.12cm}}c@{\hspace{0.12cm}}
%|c@{\hspace{0.12cm}}c@{\hspace{0.12cm}}c@{\hspace{0.12cm}}}
% {l@{\hspace{0.2cm}}|l@{\hspace{0.2cm}}|c@{\hspace{0.2cm}}c@{\hspace{0.2cm}}c@{\hspace{0.2cm}}c@{\hspace{0.2cm}}c@{\hspace{0.2cm}}}
\begin{tabular}
{@{\hspace{0.02cm}}l@{\hspace{0.08cm}}|l@{\hspace{0.08cm}}|c@{\hspace{0.08cm}}c@{\hspace{0.08cm}}c@{\hspace{0.08cm}}c@{\hspace{0.08cm}}c@{\hspace{0.08cm}}c@{\hspace{0.08cm}}c@{\hspace{0.08cm}}c@{\hspace{0.08cm}}c@{\hspace{0.08cm}}}
\hline
     & &  \multicolumn{5}{c}{PETS2009}    \\
%\hline
Semi-MVCC  & Rate   & MAE/G(0)  & G(1) & G(2)   & MSE  & NAE     \\
\hline
\multirow{3}{*}{PatchRanking \cite{liu2018leveraging}}
&5\% & 3.85   & 6.16   & 7.32   & 5.15 &0.130   \\
&10\%   & \textbf{3.56} & 6.06   & 7.17   & \textbf{4.69} & 0.126 \\
&20\%   & 3.45  & 5.70   & 7.19   & 4.52 &0.121  \\
\hline
\multirow{3}{*}{IRAST \cite{liu2020semi}}
&5\%  & 4.02   & 7.59   & 9.09   & 5.13 & 0.140 \\
&10\%  &\underline{3.60}   & 6.88   & 7.95   & \textbf{4.69} &  {0.125} \\
&20\% & 3.70   & 6.58   & 7.77   & 4.85 & 0.128    \\ 
\hline
\multirow{3}{*}{DACount \cite{lin2022semi}}
&5\% & 4.44   & 6.85   & 7.76   & 5.49 & 0.149  \\
&10\% & 4.29   & 6.75   & 7.62   & 5.37 & 0.145   \\
&20\%  &  4.07   &  5.93   & 6.80   & 5.30 & 0.149  \\
\hline
\multirow{3}{*}{CountFormer \cite{mo2024countformer}}
&5\%  &5.31 &7.11  & 8.36   &6.56  & 0.155      \\
&10\% &4.10 &6.08  &7.82    &5.24   & 0.149          \\
&20\% &3.85 & 5.52 & 7.04  &4.93  &  0.140      \\
\hline
\multirow{3}{*}{P2R \cite{lin2025p2r}}
&5\%  &  8.48   & -  & -   & 10.77  & 0.295    \\
&10\% & 7.23  &- &  -  & 8.91  &0.239    \\
&20\% &  6.78   & -  & -   & 8.63  & 0.228       \\
\hline
\multirow{3}{*}{Baseline1}
&5\%   & 4.10   & 6.29   & 7.31   & 5.41 & 0.139 \\
&10\%  & 3.88   & 6.10   & 7.28   & 5.23 & 0.135   \\
&20\%  & 3.75   & 6.12   & 7.14   & 5.02 & 0.125   \\
\hline
\multirow{3}{*}{Baseline2}
&5\%   & 3.88   & 6.17   & 7.21  & 5.08& 0.132  \\
&10\%  & 3.73   & 6.04   &6.89   & 5.02 & 0.126     \\
&20\%  & 3.70   & 6.00   & 6.65   & 4.73 & 0.125     \\
\hline
\hline
% \multirow{3}{*}{MVPR (Ours)}
% &5\% & \underline{3.78}   & \underline{5.99}   & \underline{6.88}   & \underline{4.98} & \underline{0.129} \\
% &10\% & 3.63   & \underline{5.89}   & \underline{6.25}   & \underline{4.94} & \underline{0.124} \\
% &20\% & \underline{3.44}   & \underline{5.56}   & \underline{6.48}   & \textbf{4.46} & \underline{0.121} \\
% \hline
% \multirow{3}{*}{MVUR (Ours)}
% &5\%   & \textbf{3.68}   & \textbf{5.92}   & \textbf{6.83}   & \textbf{4.88} & \textbf{0.127}   \\
% &10\%  & \underline{3.60}  &\textbf{5.41}   & \textbf{6.23}   & \underline{4.94} & \textbf{0.122}  \\
% &20\%  & \textbf{3.33}   & \textbf{5.30}   & \textbf{6.13}   & \underline{4.54} & \textbf{0.120}\\
% \hline
\multirow{3}{*}{MVPR (Vanilla)}
&5\% & \underline{3.71}   & \underline{5.99}   & \underline{6.88}   & \underline{4.94} & \textbf{0.126} \\
&10\% & 3.63   & \underline{5.44}   & \underline{6.25}   & \underline{4.94} & \underline{0.125} \\
&20\% & \textbf{3.29}   & \underline{5.56}   & \underline{6.48}   & \textbf{4.46} & \underline{0.121} \\
\hline
\multirow{3}{*}{MVUR (Ours)}
&5\%   & \textbf{3.68}   & \textbf{5.92}   & \textbf{6.83}   & \textbf{4.88} & \underline{0.127}   \\
&10\%  & \underline{3.60}  &\textbf{5.41}   & \textbf{6.23}   & \underline{4.94} & \textbf{0.122}  \\
&20\%  & \underline{3.33}   & \textbf{5.30}   & \textbf{6.13}   & \underline{4.54} & \textbf{0.120}\\
\hline
\end{tabular}
\vspace{0.1cm}
\caption{The comparison of the semi-supervised multi-view counting performance on PETS2009 on 5 metrics
(the best result is in \textbf{bold} and the second best is \underline{underlined}). 
Overall, the proposed MVUR achieves the best performance on PETS2009. 
The proposed MVPR Semi-MVCC method still achieves comparable performance to PatchRanking Semi-MVCC, and is better than IRAST, DACount Semi-MVCC, and Baselines. 
%The possible reason is each camera view in PETS2009 can cover most of the same crowd in the scene, which makes the multi-view prediction ranking constraint less effective.
}
\vspace{-0.5cm}
\label{table:PETS2009_results}
\end{table}

\begin{figure}[t]
\begin{center}
   \includegraphics[width=\linewidth]{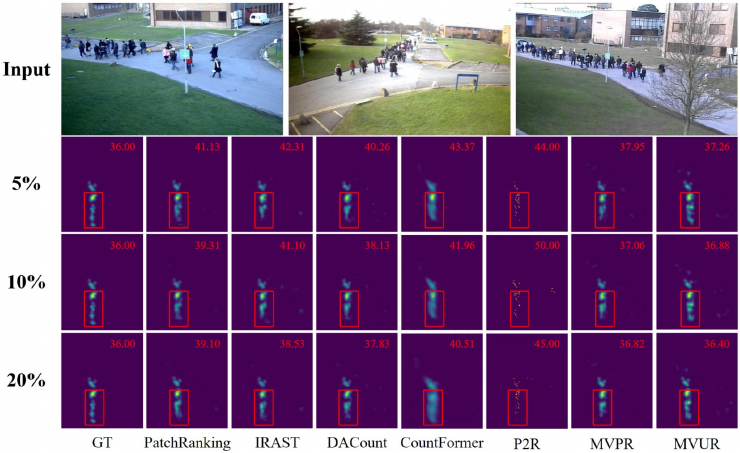}
\end{center}
%\vspace{-0.6cm}
   \caption{The visualization results of the proposed MVPR and MVUR Semi-MVCC methods and the comparison methods on PETS2009 under different annotating rates.
   % \zq{move the red boxes to the bottom parts, where our advantage is more obvious.}  %\yf{Done!}
   }
%\vspace{-0.5cm}
\label{fig:PETS2009_vis}
\end{figure}

\textbf{PETS2009.}
The performance of all methods on the PETS2009 dataset is presented in Table \ref{table:PETS2009_results}.
%(the best result is bold, and the second best is underlined). 
Generally MVUR achieves the best results and MVPR is the second among all methods. For the 5\% annotating rate, MUPR %Semi-MVCC 
performs the best on MAE/G(0), G(1), G(2), and MSE, compared to other methods. Under the 10\% and 20\%  annotating rate, MUPR %Semi-MVCC
performs the best on G(1), G(2) and NAE metrics, but both of them get the second best metrics on MAE/G(0) and MSE metrics.
These demonstrate the advantages of the proposed MVPR and MVUR over all comparison methods, and MVUR is better than MVPR due to the uncertainty ranking constraint, which avoids the overcount issue in MVPR.
% In PETS2009, View 3 covers more crowds than View 1 or 2, which is effective for the multi-view prediction ranking constraint.
% Thus, MVPR Semi-MVCC achieves the best results on the PETS2009 dataset under all annotating rates (especially on 5\% annotating rate). 
The visualization results on PETS2009 %of all methods 
are presented in Fig. \ref{fig:PETS2009_vis}, where the proposed MVUR method achieves much more accurate density predictions than comparisons, especially on the red boxes.

\begin{table}[t]
%\small
\centering
\begin{tabular}{@{\hspace{0.02cm}}l@{\hspace{0.1cm}}|c@{\hspace{0.1cm}}c@{\hspace{0.1cm}}c@{\hspace{0.1cm}}c@{\hspace{0.1cm}}c@{\hspace{0.1cm}}|c@{\hspace{0.1cm}}c@{\hspace{0.1cm}}c@{\hspace{0.1cm}}c@{\hspace{0.1cm}}c@{\hspace{0.1cm}}}
%\begin{tabular}{c@{\hspace{0.2cm}}|c@{\hspace{0.2cm}}c@{\hspace{0.2cm}}c@{\hspace{0.2cm}}c@{\hspace{0.2cm}}c@{\hspace{0.2cm}}c@{\hspace{0.2cm}}}
\hline
          &  \multicolumn{5}{c|}{PETS2009}             &  \multicolumn{5}{c}{CVCS}  \\
$\beta$   & MAE  & G(1) & G(2)   & MSE  & NAE      & MAE  & G(1) & G(2)   & MSE  & NAE   \\
\hline
0      & 3.88   & 6.15   & 7.21   & 5.08 & 0.132             & 11.75  & 17.91 & 26.12   & 15.29  & 0.105   \\
1e-4      & 3.81   & \textbf{5.66}   & \textbf{6.55}   & 5.06 & 0.129              & 10.99  & 16.77 & 24.88   & 14.06  & 0.100   \\
5e-4      & 3.85   & 6.07   & 6.87   & 5.17 & 0.131                                & 9.97  & 16.31 & 24.48   & 12.93  & 0.093   \\
1e-3     & \textbf{3.71}   & 5.99   & 6.88   & \textbf{4.94} & \textbf{0.126}     & \textbf{9.79}  & \textbf{14.97} & \textbf{23.19}   & \textbf{12.82}  & \textbf{0.090}      \\
5e-3     & 3.76   & 6.19   & 7.13   & 5.07 & 0.128                                & 10.44  & 17.08 & 24.91   & 13.43  & 0.095    \\
0.01      & 3.95   & 5.91   & 6.81   & 5.14 & 0.134                                & 10.59  & 16.95 & 25.15   & 13.87  & 0.097    \\
\hline
\end{tabular}
\vspace{0.1cm}
%\small
\caption{Ablation study of MVPR of the ranking loss weight on PETS2009 and CVCS dataset. The best result is achieved when the ranking loss weight is 0.001. We use it in all experiments.}
\vspace{-0.5cm}
\label{table:weight_ablation_study}
\end{table}

\begin{table}[t]
\centering
\begin{tabular}{cc|c@{\hspace{0.2cm}}c@{\hspace{0.2cm}}c@{\hspace{0.2cm}}c@{\hspace{0.2cm}}c@{\hspace{0.2cm}}}
\hline
\multicolumn{2}{c|}{} & \multicolumn{5}{|c}{CityStreet} \\
% \hline
 $\eta$ &$\gamma$   & MAE  & G(1) & G(2)   & MSE  & NAE       \\
\hline
0    & 1e-3  &8.20 &15.74 &30.43  &10.14   & 0.106  \\
1e-3  &0 &8.17   &13.76  &24.61  &10.35  &0.097 \\ 
1e-1   &1e-3  & 8.01   & \textbf{13.71}   & \textbf{24.43}  & 9.85 & 0.100             \\
1e-2   &1e-3  & 8.28   & 15.04   & 26.84  & 10.10 & 0.100              \\
1e-3   &1e-1  & \textbf{7.79}   & 14.21   & 25.06  & 9.76  & 0.096           \\
1e-3   &1e-2  & 8.08   & 15.29   & 29.91  & 10.00 & 0.104           \\
1e-3   &1e-3  & \textbf{7.79}   & 14.06  &  24.65   & \textbf{9.62} & \textbf{0.095}       \\
\hline
\end{tabular}
\vspace{0.1cm}
\caption{Ablation study of MVUR on the uncertainty ranking loss weight on the CityStreet dataset. The best result is achieved when both the uncertainty estimation loss and the ranking loss weight are set to 0.001, used for all experiments.}
\label{table:MVUR_weight_ablation}
\vspace{-0.5cm}
\end{table}

\begin{table}[t]
\centering
%\begin{tabular}{l@{\hspace{0.12cm}}|l@{\hspace{0.12cm}}|c@{\hspace{0.12cm}}c@{\hspace{0.12cm}}c@{\hspace{0.12cm}}|c@{\hspace{0.12cm}}c@{\hspace{0.12cm}}c@{\hspace{0.12cm}}
%|c@{\hspace{0.12cm}}c@{\hspace{0.12cm}}c@{\hspace{0.12cm}}}
\begin{tabular}{l@{\hspace{0.2cm}}|l@{\hspace{0.2cm}}|c@{\hspace{0.2cm}}c@{\hspace{0.2cm}}c@{\hspace{0.2cm}}c@{\hspace{0.2cm}}c@{\hspace{0.2cm}}}
\hline
%     & &  \multicolumn{5}{c}{PETS2009}    \\
%\hline
Training  & Rate   & MAE/G(0)  & G(1) & G(2)   & MSE  & NAE     \\
\hline
\multirow{3}{*}{RandomSelection}
&5\% & \textbf{3.71}   & \textbf{5.99}   & \textbf{6.88}   & \textbf{4.94} & \textbf{0.126}       \\
&10\% & 3.63   & 5.44   & 6.25   & 4.94 & 0.125         \\
&20\% & \textbf{3.29}   & 5.56   & 6.48   & \textbf{4.46} & 0.121       \\
\hline
\multirow{3}{*}{FixedAlternative}
&5\% & 3.86   & 6.72   & 7.46   & 5.17 & 0.137       \\
&10\% & 3.65   & 6.37   & 7.14   & 4.92 & 0.132      \\
&20\% & 3.70   & 5.29   & 6.07   & 4.89 & 0.126      \\
\hline
\multirow{3}{*}{Simultaneous}
&5\%  & 3.81   & 6.52   & 7.39   & 5.03 & 0.133      \\
&10\% & \textbf{3.56}   & \textbf{5.30}   & \textbf{6.19}   & \textbf{4.91} & \textbf{0.123}      \\
&20\% & 3.40  & \textbf{5.14}   & 6.00   & 4.52 & \textbf{0.116}     \\
\hline
\end{tabular}
\vspace{0.1cm}
\caption{Ablation study on the training approaches of MVPR Semi-MVCC on PETS2009 dataset. The best result is achieved by the Random alternative training method (RandomSelection) on 2 annotating rates 5\% and 20\%, and we use it as the training method.}
\vspace{-0.5cm}
\label{table:training_ablation_study}
\end{table}

\subsection{Ablation study}
% \yf{The ablation studies primarily focus on evaluating the impact of some hyperparameter selections, feature fusion methods, and backbone network choices, which are foundational architectural components. These factors are expected to have similar effects on both MVPR and MVUR models due to their shared framework. Therefore, using only the MVPR model for these experiments is sufficient to validate the performance influence of hyperparameter selections, feature fusion strategies, and backbone selection. This is because the key distinction between MVPR and MVUR lies in their handling of multi-view information, rather than the underlying architecture or feature fusion modules.
% Furthermore, to avoid redundancy and conserve resources, conducting ablation studies solely on the MVPR model ensures an efficient analysis of the baseline architecture's performance optimization. The specific differences and performance comparisons between MVPR and MVUR, particularly in terms of multi-view information fusion and usages, will be discussed in detail in Sec \ref{subsec:Comparison Methods}. This arrangement not only maintains logical clarity but also emphasizes the primary objectives of each experimental phase.}

\textbf{The effectiveness of MVPR and MVUR.}
We compare the results of the proposed MVPR and MVUR Semi-MVCC, and the `Baseline1' and `Baseline2' methods which share the same multi-view counting model with MVPR and MVUR but without any semi-supervised modules in Table \ref{table:CVCS_results}, \ref{table:PETS2009_results} and \ref{table:CityStreet_results}. 
%From the tables, we can conclude that 
MVPR and MVUR generally increase the performance of the baseline methods under all annotating rates on all 3 datasets, which demonstrates their effectiveness in utilizing the unlabeled data within the semi-supervised multi-view counting task. The results of the experiment validate the advantages of the model uncertainty ranking over the model prediction ranking, especially on the CityStreet dataset.

\textbf{The loss term weight in MVPR.}
To further study the influence of the multi-view prediction ranking loss on the model training, we conduct experiments by tuning the loss term weight $\beta$ in Table \ref{table:weight_ablation_study}. The experiments are conducted on the PETS2009 and CVCS datasets using the 5\% annotating rate. Table \ref{table:weight_ablation_study} shows the best result is achieved by setting the loss term weight value to 0.001. Increasing or decreasing the ranking loss weight may harm the multi-view counting model training or reduce the effect of the multi-view prediction ranking constraint, respectively. We use the fixed weight 0.001 for the ranking loss under all annotating rates.

\textbf{The loss term weight in MVUR.}
For the same reason as above, we conduct experiments to study the influence of the various losses on the model training. The experiments are only conducted on CityStreet with the 5\% annotating rate. Table \ref{table:MVUR_weight_ablation} shows that the best result is achieved by setting the values of $\eta$ and $\gamma$ to 0.001. From the results, the most suitable loss weight item should be set as $\eta=0.001$ and $\gamma=0.001$ as illustrated in the Implementation details.

% \begin{table}[t]
% %\small
% \centering
% % \begin{tabular}{@{\hspace{0.4cm}}c@{\hspace{0.4cm}}@{\hspace{0.4cm}}l@{\hspace{0.4cm}}|c@{\hspace{0.4cm}}c@{\hspace{0.4cm}}c@{\hspace{0.4cm}}c@{\hspace{0.4cm}}c@{\hspace{0.4cm}}}
% \begin{tabular}{c@{\hspace{0.2cm}}c@{\hspace{0.2cm}}|c@{\hspace{0.2cm}}c@{\hspace{0.2cm}}c@{\hspace{0.2cm}}c@{\hspace{0.2cm}}c@{\hspace{0.2cm}}}
% \hline
% &  \multicolumn{6}{c}{CityStreet}             \\
% $\eta$ &$\gamma$    & MAE  & G(1) & G(2)   & MSE  & NAE       \\
% \hline
% 1e-1   &1e-3  & 8.01   & \textbf{13.71}   & \textbf{24.63}  & 9.85 & 0.100             \\
% 1e-2   &1e-3  & 8.28   & 15.04   & 26.84  & 10.10 & 0.100              \\
% %1e-3   &1e-3  & 3.85   & 6.07   & 6.87   & 5.17 & 0.131                                 \\
% 1e-3   &1e-1  &7.79 & 14.21  & 25.06  &9.76  & 0.096           \\
% 1e-3   &1e-2  & 8.08  & 15.29   & 29.91  & 10.00 &  0.104                                \\
% 1e-3   &1e-3  & \textbf{7.77}   & 13.74   & 26.05   & \textbf{9.49} & \textbf{0.095}       \\
% \hline
% \end{tabular}
% \vspace{-0.5cm}
% %\small
% \caption{Ablation study on the ranking loss weight on CityStreet dataset. The best result is achieved when both the uncertainty estimation loss and the ranking loss weight are set to 0.001. Hence, we employ it in all experiments.}
% \vspace{-0.5cm}
% \label{table:MVUR_weight_albation}
% \end{table}

\textbf{The method of training.}
We also conduct the ablation study on the training method and take MVPR Semi-MVCC as an example because MVPR and MVUR have the same image encoder and decoder. The experiment is conducted on the PETS2009 dataset with the 5\% annotating rate.
The current method of training MVPR Semi-MVCC is as follows and we also propose other training methods for comparison.
%\begin{itemize}
\begin{itemize}
  \item \textbf{Random selection training}: This is the training method reported in the previous experiments, which trains the model by random selection---inputting the labeled and unlabeled data. Specifically, in each iteration, only 1 example is input in the model, either a labeled or unlabeled one, in a 50\% to 50\% probability. Then, the 1 labeled or unlabeled example is randomly selected from the labeled training set or the unlabeled training set, respectively.
      %Thus, the labeled and unlabeled examples are randomly and alternatively input into the model for training.
  \item \textbf{Fixed alternative training}: Compared to the random alternative training, the labeled and unlabeled examples are also alternatively input into the model, whose inputting order is fixed---input 1 randomly selected labeled example at the first iteration, and then input 1 randomly selected unlabeled example in the next iteration.
  \item \textbf{Simultaneous training}: In each iteration, 1 labeled example and 1 unlabeled example are input into the model simultaneously for training. The labeled example or the unlabeled example is also randomly selected from the labeled or the unlabeled training set, respectively.
% \end{itemize}
\end{itemize}

We perform the ablation study by training the MVPR Semi-MVCC
%semi-supervised multi-view counting model 
according to the aforementioned 3 training methods
while keeping all other settings the same. The experimental results are summarized in Table \ref{table:training_ablation_study}.
The random alternative training achieves the best results on 2 annotating rates 5\% and 20\%. The possible reason is that the random alternative training method introduces more randomness in the model with more generous abilites compared to the other 2 training methods.
We use random alternative training to train the proposed MVPR Semi-MVCC as well as MVUR Semi-MVCC in all experiments.

% \yf{
% \textbf{The impact of different fusion methods.}
% We also attempted other feature
% fusion methods on the 5% CityStreet dataset, and the results are shown in Table 9.
% Compared to the max-pooling (max) and add fusion methods, the concatenation(cat)
% method achieved the best results in all metrics except for MSE.
% }

\textbf{The impact of different fusion methods.}
We also conduct ablation studies on the impact of feature fusion methods in Table \ref{table:training_fusion_study}, using the CityStreet dataset with 5\% annotation rate for MVPR. Compared to the max-pooling and addition fusion methods, the concatenation method achieved better results according to most metrics. Thus, we use concatenation as the multi-view fusion method in the experiments.

% \textcolor{red}{
% \textbf{Impact of different feature extractors.}
% We employed the same VGG feature extractor as the state-of-the-art methods MVMSR and CVCS. Simultaneously, we also conducted experiments on the U-Net and FCN (in MVMS) feature extractors. The experimental results, as shown in Table \ref{table:training_encoder_study}, were obtained by conducting the entire process on the 5\% CityStreet dataset. In the table, ``without" indicates the absence of the MVPR semi-supervised approach, while ``with" represents its inclusion. It can be observed that our method achieved relatively better performance.
% }

\textbf{Impact of different feature extractors.}
We employ the same VGG feature extractor as the state-of-the-art methods MVMSR and CVCS. Simultaneously, we also conducted experiments on the U-Net and FCN (in MVMS) feature extractors in Table \ref{table:training_encoder_study}, using the CityStreet dataset with 5\% annotation rate for MVPR. 
%In the table, ``without" indicates the absence of the model prediction ranking constraint, while ``with" represents its inclusion. 
Our method achieves the best performance using VGG as the feature extractor. Thus, we use it in all experiments as the feature extraction backbone model. Regardless of which feature extractor is used, our MVPR approach always achieves better results than without, which also indicates the effectiveness of MVPR.

\begin{table}[t]
\centering
%\begin{tabular}{l@{\hspace{0.12cm}}|l@{\hspace{0.12cm}}|c@{\hspace{0.12cm}}c@{\hspace{0.12cm}}c@{\hspace{0.12cm}}|c@{\hspace{0.12cm}}c@{\hspace{0.12cm}}c@{\hspace{0.12cm}}
%|c@{\hspace{0.12cm}}c@{\hspace{0.12cm}}c@{\hspace{0.12cm}}}
\begin{tabular}{l@{\hspace{0.2cm}}|c@{\hspace{0.2cm}}c@{\hspace{0.2cm}}c@{\hspace{0.2cm}}c@{\hspace{0.2cm}}c@{\hspace{0.2cm}}}
\hline
Fusion    & MAE/G(0)  & G(1) & G(2)   & MSE  & NAE     \\
\hline
addition &8.86 &15.32 &28.06 &10.41 &0.112       \\
max-pooling &8.45 &15.80 &29.63 &10.16 &0.105      \\
% cat &\textbf{8.28} &\textbf{14.44} &\textbf{27.53} &10.20 &\textbf{0.099}      \\
concatenation &\textbf{7.99} &\textbf{15.24} &\textbf{26.85} &\textbf{9.96} &\textbf{0.097}      \\
\hline
\end{tabular}
\vspace{0.1cm}
\caption{Performance comparison of MVPR on CityStreet with 5\% annotation rate under three different feature fusion methods: addition, max-pooling, and concatenation. The highlighted text indicates the best performance.}
\vspace{-0.5cm}
\label{table:training_fusion_study}
\end{table}

\begin{table}[t]
\centering
%\begin{tabular}{l@{\hspace{0.12cm}}|l@{\hspace{0.12cm}}|c@{\hspace{0.12cm}}c@{\hspace{0.12cm}}c@{\hspace{0.12cm}}|c@{\hspace{0.12cm}}c@{\hspace{0.12cm}}c@{\hspace{0.12cm}}
%|c@{\hspace{0.12cm}}c@{\hspace{0.12cm}}c@{\hspace{0.12cm}}}
\begin{tabular}{l@{\hspace{0.2cm}}c@{\hspace{0.2cm}}|c@{\hspace{0.2cm}}c@{\hspace{0.2cm}}c@{\hspace{0.2cm}}c@{\hspace{0.2cm}}c@{\hspace{0.2cm}}}
\hline
%     & &  \multicolumn{5}{c}{PETS2009}    \\
%\hline
Encoder & MVPR   & MAE/G(0)  & G(1) & G(2)   & MSE  & NAE     \\
\hline
FCN7 &  &12.30 &20.74 &32.37 &14.83 &0.155      \\
FCN7 & \checkmark &10.51 &18.09 &31.46 &12.41 &0.136     \\
\hline
U-Net &  &9.96 &15.99 &28.17 &12.05 &0.126 \\
U-Net & \checkmark &8.30 &14.94 &28.11 &10.05 &0.103\\
\hline
VGG &  &10.41 &18.61 &29.40 &12.89 &0.123\\
VGG (Ours) & \checkmark  &\textbf{7.99} &\textbf{15.24} &\textbf{26.85} &\textbf{9.96} &\textbf{0.097}      \\
\hline
\end{tabular}
\vspace{0.1cm}
\caption{Performance comparison of MVPR of different feature extractors on CityStreet with 5\% annotation rate. The column MVPR indicates the inclusion/absence of the model prediction ranking constraint. The best performance is indicated in bold.}
\vspace{-0.5cm}
\label{table:training_encoder_study}
\end{table}

\begin{table}[ht]
\small
\centering
%\begin{tabular}{l@{\hspace{0.12cm}}|l@{\hspace{0.12cm}}|c@{\hspace{0.12cm}}c@{\hspace{0.12cm}}c@{\hspace{0.12cm}}|c@{\hspace{0.12cm}}c@{\hspace{0.12cm}}c@{\hspace{0.12cm}}
%|c@{\hspace{0.12cm}}c@{\hspace{0.12cm}}c@{\hspace{0.12cm}}}
\begin{tabular}{l@{\hspace{0.08cm}}l@{\hspace{0.08cm}}|c@{\hspace{0.08cm}}c@{\hspace{0.08cm}}|c@{\hspace{0.08cm}}c@{\hspace{0.08cm}}|c@{\hspace{0.08cm}}c@{\hspace{0.08cm}}}
\hline
{Dataset} &  &  \multicolumn{2}{c|}{CVCS}  &  \multicolumn{2}{c|}{CityStreet}  &  \multicolumn{2}{c}{PETS2009} \\
%\hline
{Method}   & Rate    & MAE  & NAE  & MAE  & NAE    & MAE  & NAE           \\
\hline
MVMS \cite{zhang2019wide}        &100\%    & 9.30   & 0.080   & 7.36  & 0.096 &   3.49   & 0.124   \\
MVMSR \cite{zhang2022wide}       &100\%    & -   & -   & 6.98  & 0.086 &   3.62   & 0.130   \\
3DCounting \cite{zhang2020_3d}   &100\%    & 13.30   & 0.123   & 7.54   & 0.091 & 3.15     & 0.113   \\
3DAttention \cite{zhang2022_3d}  &100\%    & -   & -   & 7.12   & - & 3.20     & -   \\
CVCS \cite{zhang2021cross}       &100\%    & \textbf{7.22}   & \textbf{0.062}   & -   & - & -     & -   \\
CVF \cite{zheng2021learning}     &100\%    & -   & -   & 7.08   & - & 3.08     & -   \\
CoCo-GCN \cite{zhai2022co}       &100\%    & -   & -   & 6.19   & 0.084 & \textbf{2.97}     & \textbf{0.109}   \\
CF-MVCC-C \cite{zhang2022calibration} &100\%  & 13.90   & 0.118   & 8.06   & 0.102 & 3.46     & 0.116   \\
\hline
\multirow{4}{*}{MVPR (Vanilla)}
&5\%    & 9.79   & 0.090  & 7.99   & 0.097  & 3.71  & 0.126    \\
&10\%   & 9.36   & 0.085  & 7.47   & 0.090  & 3.63  & 0.125    \\
&20\%   & 8.64   & 0.080  & 7.28   & 0.092  & 3.29  & 0.121    \\
% &100\%   & 7.60  & 0.070  & 6.99   & 0.081  & 3.35  & 0.113   \\
&100\%   & 8.25  & 0.074  & 6.99   & 0.081  & 3.20   & \underline{0.110}   \\
\hline
\multirow{4}{*}{MVUR (Ours)}
&5\%     & 9.73  & 0.088  & 7.79   & 0.101  & 3.68  & 0.127   \\
&10\%    & 9.01  & 0.082  & 6.51   & 0.078  & 3.60  & 0.122   \\
&20\%    & 8.09  & 0.075  & \underline{5.82}   & \underline{0.071}  & 3.33  & 0.120   \\
&100\%   & \underline{7.65}  & \underline{0.068}  &\textbf{5.57}   & \textbf{0.069} & \underline{3.05}  & \textbf{0.109}   \\ 
\hline
\end{tabular}
\vspace{0.1cm}
\caption{The performance comparison of fully-supervised methods and the proposed MVPR and MVUR semi-supervised method (the best result is in \textbf{bold} and the second best is \underline{underlined}). Overall, the proposed MVPR and MVUR methods under the 20\% annotating rate achieve comparable performance to fully-supervised methods MVMS, MVMSR, 3DCounting, 3DAttention, CVF, and CF-MVCC-C. When the MVUR is fully-supervised, it achieves the lowest NAE on the CityStreet and PETS2009 datasets.}
\vspace{-0.5cm}
\label{table:fully_supervised_comparison}
\end{table}

\subsection{Comparison with fully-supervised methods}
\label{subsec:Comparison Methods}
% \textbf{Comparison with fully-supervised methods.}
We conducted extra experiments to validate the proposed method by comparing it with fully-supervised multi-view counting methods \cite{zhang2019wide,zhang2022wide,zhang2020_3d,zhang2022_3d,zhang2021cross,zheng2021learning,zhai2022co}, which use 100\% annotations of the datasets, in Table \ref{table:fully_supervised_comparison}.
%including are all fully supervised multi-view counting methods which use 100\% annotations in the datasets. 
\cite{zhang2022calibration} is a calibration-free method with no ground-plane density map annotations.
%We can conclude that t
The proposed MVUR 
%Semi-MVCC method 
and MVPR Semi-MVCC method (with only 20\% annotated data)  already achieves comparable or even better performance to most of the state-of-the-art fully-supervised methods, such as MVMS \cite{zhang2019wide}, MVMSR \cite{zhang2022wide}, 3DCounting \cite{zhang2020_3d}, and 3DAttention \cite{zhang2022_3d}, CVF \cite{zheng2021learning} and CF-MVCC-C \cite{zhang2022calibration}, which shows that our method indeed reduces the demand for annotated data in multi-view counting tasks.

\begin{table}[t]
\small
\centering
\begin{tabular}{l@{\hspace{0.2cm}}|c@{\hspace{0.2cm}}c@{\hspace{0.2cm}}c@{\hspace{0.2cm}}c@{\hspace{0.2cm}}c@{\hspace{0.2cm}}}
\hline
loss     & MAE/G(0)  & G(1) & G(2)   & MSE  & NAE     \\
\hline
MSE& \textbf{7.99} & \textbf{15.24} & \textbf{26.85} & \textbf{9.96} & \textbf{0.097}  \\
Point-Supervision &10.31 & - & - & 12.73  &0.118      \\
\hline
\end{tabular}
\vspace{0.1cm}
\caption{Performance comparison of loss functions of MVPR on CityStreet with 5\% annotation rate under two different loss functions: MSE and point supervision. The highlighted text indicates the best performance.}
%\vspace{-0.5cm}
\label{table:comparison of loss}
\end{table}

\begin{table}[t]
% \small
\centering
\begin{tabular}
{@{\hspace{0.2cm}}l@{\hspace{0.2cm}}|c@{\hspace{0.2cm}}c@{\hspace{0.2cm}}c@{\hspace{0.2cm}}|c@{\hspace{0.2cm}}c@{\hspace{0.2cm}}c@{\hspace{0.2cm}}}
% {@{\hspace{0.2cm}}l@{\hspace{0.2cm}}|c@{\hspace{0.2cm}}c@{\hspace{0.2cm}}c@{\hspace{0.2cm}}c@{\hspace{0.2cm}}c@{\hspace{0.2cm}}|c@{\hspace{0.2cm}}c@{\hspace{0.2cm}}c@{\hspace{0.2cm}}c@{\hspace{0.2cm}}c@{\hspace{0.2cm}}}
% {l|ccccc|ccccc}
%\begin{tabular}{c@{\hspace{0.2cm}}|c@{\hspace{0.2cm}}c@{\hspace{0.2cm}}c@{\hspace{0.2cm}}c@{\hspace{0.2cm}}c@{\hspace{0.2cm}}c@{\hspace{0.2cm}}}
\hline
&  \multicolumn{3}{c|}{CityStreet}             &  \multicolumn{3}{c}{PETS2009}  \\
Method   & MAE   & MSE  & NAE      & MAE   & MSE  & NAE   \\
\hline
MVMS \cite{zhang2019wide}  &7.36	& 9.02	&0.096&\textbf{3.49}& 4.83&\textbf{0.124}\\
MVMSR \cite{zhang2022wide}   &\textbf{6.98}& 8.49	&\textbf{0.086}&3.62& 4.93	&0.130 \\
CVCS-Test \cite{zhang2021cross}    & 11.09 	&-	&0.124& 5.33 	&-	&0.174 \\
CVCS-Finetune \cite{zhang2021cross}  
&9.28 	&-	&0.112&4.12 	&-	&0.135\\
\hline
MVPR-Test  &11.78  &14.81	&0.129&5.89 &7.15	&0.199 \\
MVPR-Finetune   &8.01 &9.80	&0.098&3.65 &4.85	&0.125 \\
MVUR-Test   & 11.08 &13.97	& 0.123& 5.68 	& 6.88	&0.164\\
MVUR-Finetune  & 7.88 &\textbf{9.68} &0.098& 3.60 & \textbf{4.78}	& 0.125	\\
\hline
\end{tabular}
\vspace{0.1cm}
%\small
\caption{Generalization experiments were performed on real datasets CityStreet and PETS2009 using CVCS-pre-trained models. The first three methods are fully supervised on both datasets; MVPR/MVUR models are trained with 5\% labeled CVCS data. ``Test'' denotes direct evaluation, while ``Finetune'' means refining models with 5\% labeled target-dataset data.
}
\vspace{-0.5cm}
\label{table:training_generalization_study}
\end{table}

Furthermore, under the fully supervised training setting (100\% annotation rate), where both the density map prediction loss and the model uncertainty estimation loss are enforced on the full training set, the proposed MVUR achieves the lowest NAE on the PETS2009 dataset and gets the lowest MAE and NAE on the CityStreet dataset, outperforming several fully-supervised methods: MVMS, MVMSR, 3DCounting, 3DAttention, CVF, and CF-MVCC-C. Furthermore, the proposed MVUR (100\%) achieves close performance to the best fully-supervised SOTA methods CVCS and CoCo-GCN on the 3 datasets. Note that MVUR does not use a specially designed camera selection and fusion module as in CVCS, or the complicated graph CNN-based fusion method as in CoCo-GCN, but only uses the model uncertainty ranking constraint on a baseline multi-view crowd counting network. This demonstrates the model uncertainty ranking constraint's effectiveness for better multi-view crowd counting performance.

\subsection{Comparison of loss functions.} 
\textcolor{black}{
As demonstrated in Table \ref{table:comparison of loss}, MSE loss outperforms point-supervision loss across all metrics, mainly because our core ranking loss fails to work under point-supervision. Point-supervision only provides sparse point annotations without dense 2D density maps, lacking spatial cues to support the ranking constraint. Thus, the model under point-supervision can only learn from limited labeled lata, with no auxiliary ranking loss guidance. In Contrast, MSE loss could preserve the dense spatial structure needed for the ranking loss to leverage unlabeled data. Therefore, we choose MSE loss as the main loss function. }

\textcolor{black}{\subsection{Generalization of MVPR and MVUR}
We evaluate generalization on real datasets CityStreet and PETS2009 using models trained on synthetic CVCS (5\% labeled data). Table \ref{table:training_generalization_study} presents results: ``-Test'' denotes direct testing, ``-Finetune'' means fine-tuning with 5\% labeled real data first. ``MVMS'' \cite{zhang2019wide} and ``MVMSR'' \cite{zhang2022wide} are fully supervised methods trained on 100\% labeled data of target datasets; ``CVCS'' uses GAN-based domain adaptation for generalization from fully-labeled CVCS dataset. Though MVPR/MVUR perform modestly in direct testing, fine-tuning with 5\% real labeled data yields better generalization than ``CVCS'' and near-comparable results to fully supervised methods. This demonstrates their strong generalization: though trained with limited labels, but also adapt well to novel scenes via minimal fine-tuning data.}

\subsection{Training efficiency, computational cost, and memory usage.}
% Without using ranking branches, the total computation of the model is 0.50 TFLOPs, with a total parameter count of 7.81 M and a GPU memory usage of 11.70 GB. If ranking branches are used, the total computational cost of the model is 0.81 TFLOPs, the total parameter count is 18.50 M, and the GPU memory usage is 12.40 GB. That is to say, the ranking branches will bring an additional 0.31 TFLOPs calculation consumption, 10.69 million parameter count, and 0.70 G storage consumption. 
We present the training efficiency, computation cost, and memory usage of different methods in Table \ref{table:flops}, which is conducted on CVCS dataset.
The proposed MVPR and MVUR have similar FLOPs, model parameters (Params), GPU memory (Mem) usage, and training time (Time) to other comparison methods. The reason is they share the same multi-view crowd counting model (MVCC), which accounts for most of the computation resources. MVUR needs more parameters, memory, and training time than MVPR due to the extra model uncertainty decoder. Overall, compared to the original MVCC model, the extra computation cost caused by the model prediction and model uncertainty ranking branches is not significant.

\begin{table}[t]
\centering
\small
% \begin{tabular}{l@{\hspace{0.12cm}}|l@{\hspace{0.12cm}}|c@{\hspace{0.12cm}}c@{\hspace{0.12cm}}c@{\hspace{0.12cm}}|c@{\hspace{0.12cm}}c@{\hspace{0.12cm}}c@{\hspace{0.12cm}}
% |c@{\hspace{0.12cm}}c@{\hspace{0.12cm}}c@{\hspace{0.12cm}}}
\begin{tabular}{l@{\hspace{0.12cm}}|c@{\hspace{0.12cm}}c@{\hspace{0.12cm}}c@{\hspace{0.12cm}}c@{\hspace{0.12cm}}}
% \begin{tabular}{l|cccc}
\hline
%\hline
 Method  & FLOPs/G  & Params/MB & Mem/GB & Time/h\\
\hline
PatchRanking \cite{liu2018leveraging} &1882.03 & 152.66   & 10.05 &51.70 \\
IRAST \cite{liu2020semi} &  1882.03  & 152.66  & 9.61 &129.70  \\
DACount \cite{lin2022semi} & 1879.49   &  152.90  & 10.03& 68.42  \\
CountFormer \cite{mo2024countformer} & 1527.24 & 75.99 & 9.88 & 83.24 \\
P2R \cite{lin2025p2r} &709.05 &71.23 &34.19 &94.64 \\
Baseline1 & 1928.94   & 150.08    &  9.97 & 66.77  \\
Baseline2 & 1928.94   & 150.08   & 9.97 &75.41 \\
\hline
MVPR (Vanilla) & 1928.94 & 150.08  & 9.98 & 59.87 \\
MVUR (Ours) & 2195.19   & 150.45 & 10.31  & 67.58\\
\hline
\end{tabular}
\vspace{0.1cm}
\caption{Consumption comparison on CVCS: training efficiency, computational cost, memory usage, and the training time cost.}
% \zq{add other methods and reference.}}
\vspace{-0.5cm}
\label{table:flops}
\end{table}

\section{Discussion and Conclusion}
In this paper, we focus on reducing the demand for labeled data in the multi-view crowd counting task and propose two semi-supervised multi-view counting frameworks: 
a vanilla model--multi-view model prediction ranking (MVPR), and our complete model--multi-view model uncertainty ranking (MVUR) semi-MVCC methods. As far as we know, this is the first semi-supervised study in the multi-view crowd counting area.
MVPR ranks the predictions of models with different numbers of input views as an extra constraint, and MVUR estimates the model uncertainties for different numbers of input views and adds a corresponding ranking order constraint. MVUR avoids the over-count issue of MVPR, and the limitation that MVPR is ineffective on scenes where each view can cover most of the crowd.
The experiments demonstrate the effectiveness of the proposed MVPR and MVUR Semi-MVCC methods compared with other semi-supervised methods originally designed for single-image counting tasks. In the future, the proposed multi-view ranking constraint can be extended to other multi-view vision tasks, such as semi-supervised multi-view people detection or tracking, which can reduce the annotation cost for multi-view images for more practical application scenarios.
%\section*{Declarations}
\textbf{Ethical consideration}.  We use public synthetic and real multi-view image datasets for researching semi-supervised multi-view counting. The proposed semi-supervised methods will reduce the demand for annotated multi-view images.
% \textbf{Data availability}. The data supporting this study's findings is available from the public datasets, with sources referred to in the paper. 

\section*{Acknowledgements}
This work was supported in parts by NSFC (62202312), Guangdong Basic and Applied Basic Research Foundation (2023B1515120026), Shenzhen Science and Technology Program (KQTD 20210811090044003, RCJC20200714114435012), and Scientific Foundation for Youth Scholars and Scientific Development Funds of Shenzhen University.
 
% \vfill
\newpage
% \clearpage
\bibliographystyle{IEEEtran}
\bibliography{ref}

\end{document}